\documentclass{article}

 \usepackage[preprint]{neurips_2026}

\usepackage[utf8]{inputenc} 
\usepackage[T1]{fontenc}    
\usepackage{hyperref}       
\usepackage{url}            
\usepackage{booktabs}       
\usepackage{amsfonts}       
\usepackage{nicefrac}       
\usepackage{microtype}      
\usepackage{xcolor}         

\usepackage{multirow}
\usepackage{graphicx}
\usepackage[dvipsnames]{xcolor}
\usepackage[table]{xcolor}

\usepackage{amsmath} 
\usepackage{amsthm} 
\usepackage{amssymb} 
\usepackage{mathtools}
\usepackage{algorithm}
\usepackage{algpseudocode}
\usepackage{caption}
\usepackage{wrapfig}

\usepackage{enumitem}

\usepackage{comment}

\usepackage[svgnames]{xcolor}
\usepackage{framed}
\definecolor{shadecolor}{named}{WhiteSmoke}

\definecolor{maincolor}{HTML}{4D6FA3}
\definecolor{emphtextcolor}{HTML}{FFFFFF}

\usepackage{colortbl}
\title{ReCache: Learning Budget-Aware Caching Schedules for Diffusion Models via REINFORCE}

\author{\textbf{Mishan Aliev}\thanks{Equal contribution. Correspondence: Mishan Aliev <maliev@hse.ru>, Eva Neudachina <eneudachina@hse.ru>, Ilya Bykov <philurame@gmail.com>.}\hspace{0.15cm}$^{,1}$, \textbf{Eva Neudachina}$^{*,1}$, \textbf{Ilya Bykov}$^{*,1}$, \textbf{Aleksandr Oganov}$^1$, \\ \textbf{Kirill Struminsky}$^{2,1}$,  \textbf{Aibek Alanov}$^1$, \textbf{Denis Rakitin}$^1$ \vspace{.5em}\\
\begin{tabular}{ll}
$^1$HSE University, Russia & \\
$^2$Yandex Research, Russia & \\
\end{tabular}
}

\begin{document}

\newcommand{\bx}{\mathbf{x}}
\newcommand{\bw}{\mathbf{w}}
\newcommand{\bz}{\mathbf{z}}
\newcommand{\by}{\mathbf{y}}
\newcommand{\bc}{\mathbf{c}}
\newcommand{\bb}{\mathbf{b}}
\newcommand{\ba}{\mathbf{a}}
\newcommand{\bs}{\mathbf{s}}

\newcommand{\bA}{\mathbf{A}}
\newcommand{\bB}{\mathbf{B}}
\newcommand{\bC}{\mathbf{C}}
\newcommand{\bD}{\mathbf{D}}
\newcommand{\bS}{\mathbf{S}}
\newcommand{\bP}{\mathbf{P}}
\newcommand{\bU}{\mathbf{U}}
\newcommand{\bQ}{\mathbf{Q}}
\newcommand{\bV}{\mathbf{V}}
\newcommand{\bI}{\mathbf{I}}
\newcommand{\bF}{\mathbf{F}}
\newcommand{\bG}{\mathbf{G}}
\newcommand{\bL}{\mathbf{L}}

\newcommand{\bSigma}{\boldsymbol{\Sigma}}
\newcommand{\bmu}{{\boldsymbol{\mu}}}
\newcommand{\bzero}{\mathbf{0}}

\newcommand{\bX}{\mathbf{X}}
\newcommand{\bY}{\mathbf{Y}}
\newcommand{\dd}{\mathrm{d}}
\newcommand{\score}{\nabla \log}
\newcommand{\scorex}{\nabla_{\bx} \log}
\newcommand{\scorext}{\nabla_{\bx_t} \log}
\newcommand{\dist}{\mathbf{d}}

\newcommand{\R}{\mathbb{R}}
\newcommand{\E}{\mathbb{E}}

\newcommand{\calA}{\mathcal{A}}
\newcommand{\calG}{\mathcal{G}}
\newcommand{\calX}{\mathcal{X}}
\newcommand{\calP}{\mathcal{P}}
\newcommand{\calC}{\mathcal{C}}
\newcommand{\calK}{\mathcal{K}}
\newcommand{\calN}{\mathcal{N}}
\newcommand{\diag}[1]{\mathrm{diag}(#1)}
\newcommand{\pd}{p_{\text{data}}}
\newcommand{\eps}{\varepsilon}
\newcommand{\beps}{\boldsymbol{\eps}}
\renewcommand{\v}{\boldsymbol{v}}
\newcommand{\xcache}{\bx_{\text{cache}}}
\newcommand{\xfull}{\bx_{\text{full}}}

\newcommand{\structset}{\mathcal{S}}
\newcommand{\struct}{\mathbf{s}}

\newcommand{\kl}[2]{D_{\mathrm{KL}}\!\left(#1 ~ \| ~ #2\right)}
\newcommand{\normal}[3]{\mathcal{N}\left(#1 \ | \ #2, #3\right)}

\renewcommand{\algorithmicrequire}{\textbf{Input:}}
\renewcommand{\algorithmicensure}{\textbf{Output:}}

\theoremstyle{definition}
\newtheorem{lemma}{Lemma}
\newtheorem{definition}{Definition}

\newcommand{\entropyreg}{\lambda_{\text{entropy}}}
\newcommand{\losscoef}{\alpha_{\text{iq}}}

\newcommand{\loss}{\mathcal{L}}
\newcommand{\lossimg}{\mathcal{L}_{\text{image}}}
\newcommand{\lossreg}{\mathcal{L}_{\text{reg}}}
\newcommand{\meanlossreg}{\overline{\mathcal{L}_{\text{reg}}}}

\maketitle

\begin{abstract}
Modern diffusion models generate high-quality images and videos, but their iterative denoising process makes inference expensive.
Feature caching accelerates sampling by reusing or predicting intermediate activations across neighboring denoising steps, exploiting the redundancy of computations along the reverse trajectory.
In this work, we focus on the caching schedule: selecting which denoising steps should be fully recomputed.
Existing schedules are either fixed (e.g.\ uniform) or chosen adaptively from per-step error heuristics; in both cases, the actual compute cost is a side-effect of hand-tuned thresholds rather than a quantity the user can specify.
We propose \textbf{ReCache}, which inverts this: given a target budget $k$, it learns the recomputation schedule that maximizes generation quality, turning compute into a directly controllable input.
ReCache trains via policy gradients, sidestepping backpropagation through full diffusion inference, and uses no labelled data.
Generations from uncached inference serve as matching targets, paired with a reward for generation quality.
ReCache is compatible with any caching mechanism, including feature reuse and feature forecasting; for each mechanism, a single trained policy adapts across computational budgets at inference time.
ReCache consistently outperforms scheduling baselines: under a $\times5.04$ FLOPs reduction on FLUX, it reduces LPIPS by 31\% (from 0.456 to 0.316) compared to DiCache; on Wan 2.1 at a $\sim \times2.6$ speedup, it drops LPIPS by 65\% (from 0.480 to 0.169) and boosts the VBench score by 7\% (5.6 points, from 70.4 to 76.0) over uniform HiCache.
Code is available at \href{https://github.com/thecrazymage/ReCache}{\texttt{https://github.com/thecrazymage/ReCache}}.
\end{abstract}

\begin{figure}[!h]
    \centering
    \includegraphics[width=1\textwidth]{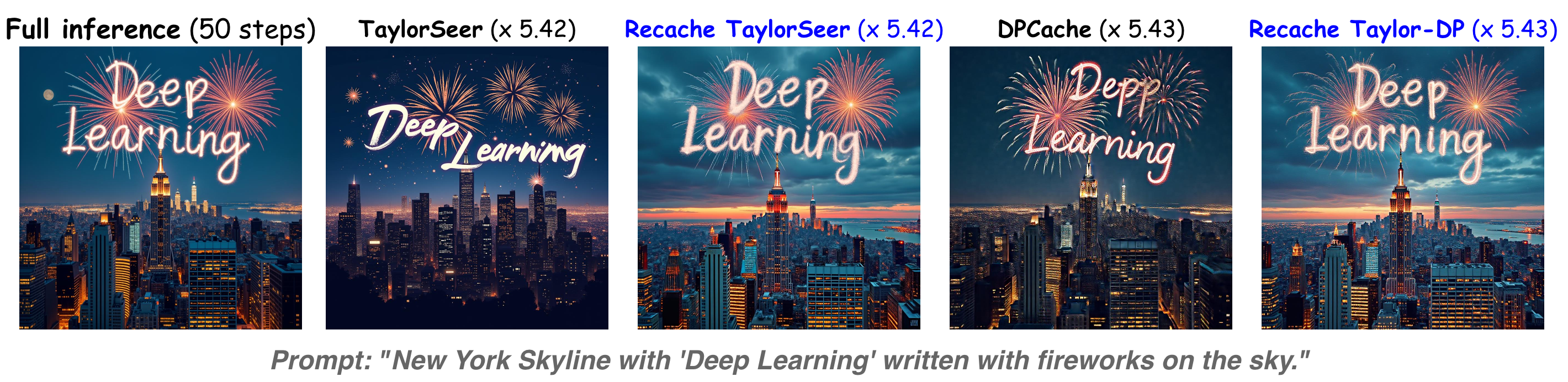}
    \caption{
        ReCache preserves text and visual fidelity under high acceleration on FLUX.1-dev~\cite{blackforest2024flux}. Under a $5.4\times$ acceleration, heuristic baselines (TaylorSeer~\cite{liu2025taylorseer}, DPCache~\cite{cui2026dpcache}) fail to maintain image quality and corrupt text spelling. By framing scheduling as an RL problem, ReCache allocates computations to the most critical steps, closely matching Full inference at the same computational cost.
    }
    \label{fig:teaser}
\end{figure}

\section{Introduction}

Diffusion models have become one of the dominant paradigms for high-fidelity image and video generation~\cite{ho2020denoising, rombach2022high, blattmann2023stable}.
Recent transformer-based architectures, such as DiT~\cite{peebles2023scalable}, have further improved scalability and generation quality, enabling increasingly powerful text-to-image and text-to-video systems.
However, these advances come at a substantial inference cost: generating a single sample requires repeatedly evaluating a large denoising network over many sampling steps.
As model size, resolution, and video length continue to grow, reducing the cost of diffusion inference has become an important practical challenge.

A prominent line of work accelerates diffusion sampling through \emph{feature caching}.
The key observation is that consecutive denoising steps often perform highly redundant computations: intermediate activations change smoothly along the reverse trajectory and can therefore be reused or approximated across nearby timesteps~\cite{ma2024deepcache, selvaraju2024fora}.
Existing caching methods typically differ along two axes: \emph{what} to cache and \emph{when} to cache. 
While some approaches reuse selected model components according to a fixed schedule~\cite{ma2024deepcache, selvaraju2024fora, chen2024delta}, others determine caching steps dynamically using error-based heuristics \cite{liu2025timestep}.
Furthermore, recent methods go beyond direct reuse by forecasting future features~\cite{liu2025taylorseer, feng2025hicache}.
Despite these differences, most methods still rely on hand-crafted scheduling rules to decide at which denoising steps the expensive full computation should be performed.

In this work, we focus on the scheduling problem, arguing that caching schedules should be learned rather than hand-crafted. 
Existing approaches, including both uniform spacing and adaptive heuristics, often aim to keep the accelerated sampling process close to the original full-inference trajectory at intermediate denoising steps. 
However, this objective can be suboptimal under a fixed computational budget: different denoising timesteps contribute unequally to the final generation, and the steps that best preserve intermediate \textit{trajectory consistency} may not coincide with those most important for the final output. 
This suggests that schedule design should not only minimize local approximation errors along the denoising path, but should also account for how each recomputation decision affects the final sample. 
We therefore argue that caching schedules should be optimized directly for \textit{output consistency}, i.e., for preserving the perceptual and semantic quality of the generated sample relative to full inference.

To achieve this, we frame caching-schedule selection as a Reinforcement Learning (RL) problem.
We propose \textbf{ReCache}, a budget-aware method that learns a scheduling policy for diffusion feature caching. 
Once a policy is trained for a specific model and caching setup, ReCache takes a target budget and predicts a schedule, determining exactly where full computation is necessary and where cached features are sufficient.
The policy is optimized via a reward with two objectives: minimizing the distance to the full-inference samples and preserving the perceptual and semantic quality of the generated images.
As shown in Figure~\ref{fig:teaser}, for a fixed computational budget, our learned policy produces outputs that more closely match the full-inference generations than those from uniform or heuristic-adaptive baselines.

We evaluate ReCache on FLUX.1-dev~\cite{blackforest2024flux}, HunyuanVideo~\cite{kong2024hunyuanvideo}, and Wan2.1~\cite{wan2025wan}, and combine it with recent caching mechanisms such as TaylorSeer~\cite{liu2025taylorseer} and HiCache~\cite{feng2025hicache}.
A single trained policy (for a given caching mechanism) adapts to multiple FLOPs budgets at inference time, avoiding the need to design a separate schedule for each acceleration regime.
Across all evaluated settings, ReCache consistently outperforms hand-crafted scheduling baselines at the same computational cost, demonstrating that learning budget-aware schedules is a simple and effective way to improve diffusion caching methods.

In summary, our main contributions are:
\vspace{-0.3em}
\begin{itemize}[leftmargin=*, itemsep=0.2em, topsep=0.2em]
    \item We frame caching schedule selection as a budget-aware RL problem that directly optimizes output consistency under a fixed compute budget, instead of relying on hand-crafted trajectory-consistency heuristics.
    \item We propose \textbf{ReCache}, a lightweight scheduling method compatible with diverse diffusion models and caching mechanisms.
    \item We show that ReCache consistently improves over hand-crafted scheduling baselines across multiple budgets, caching strategies, and state-of-the-art image and video models, including FLUX, HunyuanVideo, and Wan2.1.
\end{itemize}
\vspace{-0.3em}
\section{Preliminaries}

Diffusion and Flow Matching models~\cite{sohl2015deep, ho2020denoising, song2020score, lipman2022flow, liu2022flow, albergo2022building} generate samples by numerically integrating a learned ODE that transports a Gaussian latent $\bz \sim \calN(0, I)$ to a data sample.
A diffusion model $G$ produces a sample in $N$ discrete inference steps, determined by a timestep schedule and an ODE solver.
Each step evaluates a learned velocity $\v(\bx_t, t)$ parameterized as a neural network, and this evaluation dominates the per-step cost.
In modern image and video models, this network is a diffusion transformer~\cite{peebles2023scalable, esser2024scaling, blackforest2024flux, wan2025wan, kong2024hunyuanvideo}, whose forward pass through stacked attention and feedforward blocks is what caching methods aim to amortize.
Since our method is agnostic to the specific timestep schedule, we index inference steps by their position $\{1, 2, \ldots, N\}$ throughout the paper.

\paragraph{Feature caching.}
Recent works observe that the intermediate activations of $\v_\theta$ change slowly between adjacent inference steps and propose to amortize the per-step cost by reusing them.
The inference steps are split into two groups: \emph{cache steps}, where the network is evaluated in full and selected activations are stored, and \emph{reuse steps}, where the missing activations are reconstructed from the cache rather than recomputed. A caching schedule $\struct$ is a subset $\struct \subseteq \{1, \ldots, N\}$ of cache steps, while its size $|\struct| = k$ is a \textit{caching budget} --- it controls the number of full model evaluations in the generation process.

Apart from a problem of choosing a caching schedule, \emph{caching mechanisms} $\mathbf M$ set a rule of how the reuse steps reconstruct activations that a full forward trajectory would have produced. \emph{Direct reuse} mechanisms copy cached activations from the most recent cache step, either as block outputs (FORA~\cite{selvaraju2024fora}) or as residuals added to the current step's input ($\Delta$-DiT~\cite{chen2024delta}).
\emph{Feature forecasting} mechanisms instead extrapolate the current activations from values cached at several previous cache steps: TaylorSeer~\cite{liu2025taylorseer} fits a Taylor polynomial to the trajectory of cached features, and HiCache~\cite{feng2025hicache} replaces the Taylor basis with one better suited to diffusion feature trajectories.
Our experiments evaluate ReCache on top of both direct and forecasting methods, but ReCache is agnostic to the choice of $\mathbf{M}$.

\paragraph{Caching schedules.}
Existing methods predominantly propose policies of choosing caching schedules $\struct$ via hand-crafted rules.
Many works place cache steps at uniform intervals along the inference trajectory~\cite{chen2024delta, selvaraju2024fora, liu2025taylorseer, feng2025hicache}; others select them adaptively per generation, e.g.\ by triggering a cache step whenever the activation error exceeds a threshold~\cite{liu2025timestep, bu2025dicache}.
Both families either fix the budget implicitly or expose it only through threshold hyperparameters, motivating the budget-aware learned policy we develop in Section~\ref{sec:Method}.

An extended discussion of these prior approaches and broader related work is provided in Appendix~\ref{app:related_works}.

\section{Method}
\label{sec:Method}

\begin{figure}[!t]
    \centering
    \includegraphics[width=1\textwidth]{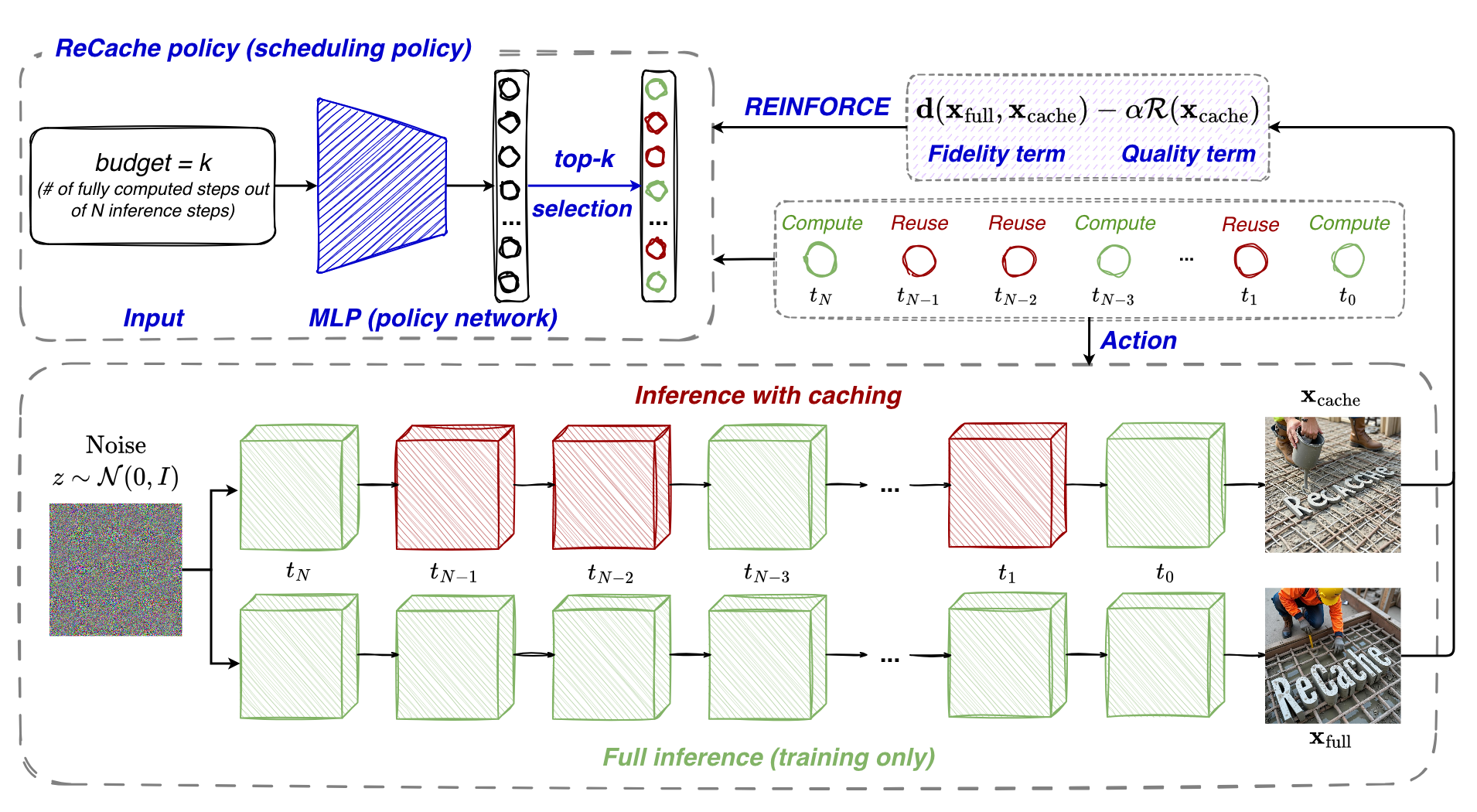}
    \caption{
        Overview of the ReCache method. Given a target computational budget $k$, a lightweight MLP policy network predicts importance logits to perform a top-$k$ selection of inference steps for full computation. During the inference with caching, the selected steps are fully computed (green blocks), while the intermediate steps reuse or forecast features (red blocks). The policy is trained using the REINFORCE algorithm without backpropagating through the diffusion process. The reward formulation balances a Fidelity term (minimizing the distance to the $\xfull$ output from the full inference) and a Quality term (maximizing the perceptual quality of the generated $\xcache$).
    }
    \label{fig:method_scheme}
\end{figure}

We introduce \textbf{ReCache}, a framework that trains a schedule policy: given a target budget $k$, the policy selects $k$ inference steps to recompute (Figure~\ref{fig:method_scheme}). We cast the training as a stochastic policy-optimization problem and parameterize the policy as a budget-conditioned distribution over $k$-subsets.

\subsection{Policy Optimization}

To evaluate a schedule's quality, we combine two complementary signals. The first signal asks that the cached generation stay close to the full-inference one. This target provides strong supervision, since the full-inference output is a direct reference for what the model would have produced without caching. Under aggressive caching, however, exact agreement with the full-inference trajectory is difficult to achieve, and the policy may need to settle for an approximation. The second signal is a reward model~\cite{xu2023imagereward, wu2023human, kirstain2023pick} that scores generation quality directly, letting the policy fine-tune perceptual details that the distance term alone cannot recover.

Concretely, let $\bz \sim \calN(0, I)$ and $\xfull = G(\bz)$, and denote by $\xcache = G(\bz \mid \struct, \mathbf{M})$ the output produced under schedule $\struct$ and caching mechanism $\mathbf{M}$. Given a distance $\mathbf{d}$ and a reward model $\mathcal{R}$,
\begin{equation}
    \lossimg(\struct) = \mathbf{d}\left(\xcache,\, \xfull\right) - \losscoef\, \mathcal{R}\left(\xcache\right).
\end{equation}

Direct minimization of $\lossimg$ over schedules is infeasible: the loss is non-differentiable in $\struct$, and the search space contains $\binom{N}{k}$ subsets. We instead lift the problem to a stochastic policy and train a distribution $p_\theta$ over $k$-subsets to minimize the expected loss
\begin{equation}
    \min_\theta\; \E_{\struct \sim p_\theta}\, \lossimg(\struct).
\end{equation}

Without further care, the policy can collapse to a deterministic schedule that is locally good but globally suboptimal. We counter this with an entropy bonus $-\entropyreg H(\theta)$ that keeps the policy exploring during training, yielding the regularized loss
\begin{equation}
    \lossreg(\struct) = \lossimg(\struct) + \entropyreg \log p_\theta(\struct).
\end{equation}

The regularized expectation $\E_{p_\theta}\, \lossreg(\struct)$ can be optimized over $\theta$ with policy-gradient methods. We use REINFORCE~\cite{williams1992simple} with a leave-one-out (LOO) baseline~\cite{kool2019buy, richter2020vargrad} for variance reduction: given $n$ i.i.d.\ samples $\struct^{(1)}, \ldots, \struct^{(n)} \sim p_\theta$ and their mean $\meanlossreg = \tfrac{1}{n}\sum_i \lossreg(\struct^{(i)})$,
\begin{equation}
    \nabla_\theta\, \E_{p_\theta(\struct)}\, \lossreg(\struct) \approx \frac{1}{n-1} \sum_{i=1}^{n} \left( \lossreg(\struct^{(i)}) - \meanlossreg \right) \nabla_\theta \log p_\theta(\struct^{(i)}).
\label{eq:reinforce-loo}
\end{equation}

\subsection{Budget-conditioned Policy Over \texorpdfstring{$k$}{TEXT}-subsets}

The policy must select a $k$-subset $\struct \subseteq \{1, \ldots, N\}$. We parameterize it with the Plackett-Luce distribution~\cite{plackett1975analysis}: each step $i$ is assigned an importance logit $\theta_i$, and a sample is drawn by sampling without replacement from the corresponding categorical distribution $k$ times. Applying the chain rule yields the closed-form log-probability
\begin{equation}
    \log p_\theta(\struct) = \sum\limits_{i=1}^k \left[ \theta_{\struct_i} - \log \sum_{j \notin \{\struct_1, \ldots, \struct_{i-1}\}} \exp(\theta_j) \right].
\label{eq:logprob}
\end{equation}

The relative importance of each step may in principle depend on the budget. At small $k$, only the most critical steps survive, and the policy must concentrate its mass on those steps. At larger $k$, the policy has more room to spread mass across the trajectory. We do not assume that the optimal step ranking is budget-invariant: we condition the logits on $k$ through a lightweight MLP with parameters $\phi$, $\theta = \text{MLP}_\phi(k)$, and train a single budget-adaptive policy by sampling $k \sim q(k)$ at every training step, so that one trained model covers a range of inference budgets without retraining. Empirically (Section~\ref{sec:ablations}), the trained policy produces \emph{nested} schedules across budgets: the $(k{+}1)$-schedule extends the $k$-schedule by a single step rather than reordering its existing selections. This suggests that under the caching objective the underlying step-importance ranking is largely budget-invariant. The MLP parameterization makes no architectural commitment to this; nestedness is discovered, not imposed, and a single set of $k$-independent logits would in retrospect suffice.

In practice we never sample sequentially. The Gumbel-Top-$k$ trick~\cite{kool2019stochastic, gadetsky2020low, struminsky2021leveraging} produces a Plackett-Luce sample in one shot by perturbing each logit with independent Gumbel noise and returning the top-$k$ indices:
\begin{equation}
    \begin{aligned}
        \mathbf{g}_i &= \theta_i - \log(-\log \mathbf{u}_i), \quad \mathbf{u}_i \overset{\text{i.i.d.}}{\sim} \mathcal{U}(0,1), \\
        \struct &= (\struct_1, \dots, \struct_k) = \operatorname*{argtop k}(\mathbf{g}_1, \dots, \mathbf{g}_N).
    \end{aligned}
\label{eq:gumbel_sampling}
\end{equation}
At inference we drop the noise and use the deterministic schedule $\struct = \operatorname*{argtop k}(\theta_1, \ldots, \theta_N)$.

\subsection{Training Algorithm}

Algorithm~\ref{alg:recache_training} summarizes ReCache training. Before training begins, we pre-sample a small dataset of $(\bx_1, \xfull)$ pairs, where each pair consists of an initial noise sample and the corresponding full-inference output of $G$. At each training step, we draw a target budget $k \sim q(k)$, obtain logits $\theta = \text{MLP}_\phi(k)$, sample $n$ schedules from the resulting Plackett-Luce distribution via the Gumbel-Top-$k$ trick, and update $\phi$ with the LOO estimator (Eq.~\ref{eq:reinforce-loo}) of the regularized loss $\lossreg$.

\begin{algorithm}
\caption{ReCache schedule policy \textbf{training}}
\begin{algorithmic}
\Require diffusion model $G$ with $N$ backbone steps, caching mechanism $\mathbf{M}$, number of leave-one-out samples $n$, distance function $\mathbf{d}$, reward function $\mathcal{R}$, loss coefficient $\losscoef$, regularizer coefficient $\entropyreg$;

\Repeat
\ForAll{elements in batch}

    \State $\bz \sim \calN(0, I)$, $\xfull = G(\bz)$
    \Comment{pre-sampled noise and full-inference image}

    \State $k \sim q(k)$
    \Comment{sample caching budget}
    \State $\theta = \text{MLP}_\phi(k)$
    \Comment{logits parameterized by MLP}

    \For{$i = 1$ to $n$}
        \State Sample $\struct^{(i)} = (\struct_1^{(i)}, \ldots, \struct_k^{(i)}) \sim p_\theta$
        \Comment{via Eq.~\ref{eq:gumbel_sampling}}

        \State $\xcache^{(i)} = G(\bz \mid \struct^{(i)}, \mathbf{M})$
        \Comment{generate image with cached inference}

        \State $\lossimg(\struct^{(i)}) = \mathbf{d}\big(\xcache^{(i)}, \xfull\big) - \losscoef\, \mathcal{R}(\xcache^{(i)})$

        \State Compute $\log p_\theta(\struct^{(i)})$
        \Comment{via Eq.~\ref{eq:logprob}}

        \State $\lossreg(\struct^{(i)}) = \lossimg(\struct^{(i)}) + \entropyreg \log p_\theta(\struct^{(i)})$
    \EndFor

    \State $\meanlossreg = \tfrac{1}{n} \sum_{i=1}^n \lossreg(\struct^{(i)})$
    \State $\loss(\struct^{(i)}) = \big(\lossreg(\struct^{(i)}) - \meanlossreg\big)\text{.stopgrad()}$
    \State $\nabla_\theta \loss(\struct) = \tfrac{1}{n-1} \sum_{i=1}^{n} \loss(\struct^{(i)})\, \nabla_\theta \log p_\theta(\struct^{(i)})$

\EndFor

\State Accumulate gradients across batch elements and update $\phi$ via backprop through $\theta = \text{MLP}_\phi(k)$.

\Until{converged}

\end{algorithmic}
\label{alg:recache_training}
\end{algorithm}

\section{Experiments}
\label{sec:Experiments}

\subsection{Experimental Setup}

We evaluate ReCache on three widely used image and video generation backbones: FLUX.1-dev~\cite{blackforest2024flux} for text-to-image generation, and HunyuanVideo~\cite{kong2024hunyuanvideo} and Wan2.1~\cite{wan2025wan} for text-to-video generation. For FLUX.1-dev, we use $N=50$ inference steps and generate images at $1024\times1024$ resolution. HunyuanVideo uses $N=50$ steps and generates videos at $480 \times 640 \times 65$ resolution, while Wan2.1 uses $N=25$ steps and $480 \times 832 \times 81$.

\paragraph{Baselines.}
ReCache is a scheduling policy and is agnostic to the underlying caching mechanism. We therefore evaluate it across several computational budgets---7, 9, and 13 full-computation steps---and combine it with state-of-the-art caching mechanisms. For feature-forecasting methods, including TaylorSeer~\cite{liu2025taylorseer} and HiCache~\cite{feng2025hicache}, we compare ReCache with a uniform schedule. DiCache~\cite{bu2025dicache} introduces Dynamic Cache Trajectory Alignment as a caching mechanism and Online Probe Profiling as a scheduling strategy. DPCache~\cite{cui2026dpcache} focuses on schedule selection and uses a modified TaylorSeer mechanism called Taylor-DP. For fair comparison with DiCache and DPCache, we keep the corresponding caching mechanism fixed and replace only the schedule with ReCache. We also include a naive step-reduction baseline, where the model is run with fewer denoising steps and no caching. All results are grouped by caching mechanism and budget.

\paragraph{Evaluation metrics.}
For text-to-image generation, we generate images for 1000 MS-COCO~\cite{lin2014microsoft} prompts and report LPIPS~\cite{zhang2018perceptual} and HPSv2~\cite{wu2023human}. LPIPS is computed against the corresponding full-inference output. We also report ImageReward~\cite{xu2023imagereward} on 200 DrawBench~\cite{saharia2022photorealistic} prompts (IR DB). For text-to-video generation, we report frame-level LPIPS and HPSv2 on 110 VBench~\cite{huang2024vbench} prompts, averaged over all frames and prompts. We additionally report the VBench score on the full VBench benchmark with 944 prompts. For all experiments, we report TFLOPs and acceleration rate to measure inference efficiency.

\paragraph{Training datasets and details.}
Each training sample contains a prompt, an initial noise sample, and the corresponding full-trajectory output. For text-to-image experiments, prompts are sampled from MS-COCO. For text-to-video experiments, we use stratified sampling over VBench dimensions and generate the corresponding full-inference videos. Each training set contains 64 samples. We use patch-wise LPIPS (PLPIPS) as the distance $\mathbf{d}$ and HPSv2 as the generation-quality reward $\mathcal{R}$. The entropy coefficient is annealed to zero during training. For FLUX.1-dev, we set $\losscoef = 7$, initialize $\entropyreg = 0.005$, and train the logit predictor for 50 epochs. For video models, we set $\losscoef = 7$, initialize $\entropyreg = 0.02$, and train the logit predictor for 120 epochs. To make ReCache budget-aware, we randomly sample the number of full-computation steps for each training sample, with most values concentrated in the range $k \in [6, 15]$.

More details about model configurations, baseline reproduction, training datasets, and hyperparameters are provided in Appendix~\ref{app:details}.

\subsection{Main experimental results}
Our experiments show that ReCache can obtain a more optimal caching schedule for any underlying caching mechanism for both image and video generation. The results show that it generally improves both LPIPS fidelity and generative quality considering HPS and VBench scores, which were specifically constructed to be aligned with human perception. The improvements are especially pronounced under tight compute budgets.

\begin{figure}[t]
    \centering

    \begin{minipage}[t]{0.69\linewidth}
        \vspace{0pt}
        \centering
        \captionof{table}{ReCache caching schedule policy for FLUX.1-dev. Comparison across caching mechanisms and inference budgets k. In most settings, ReCache achieves better full-inference trajectory fidelity and higher image quality under the same computational budget, with particularly strong gains in low-compute regimes. $O$ denotes the forecasting polynomial order.}
        
        \resizebox{\linewidth}{!}{
        \begin{tabular}{l c c c c c}
\toprule
\multirow{2}{*}{} &
\multicolumn{2}{c}{\textbf{Acceleration}} &
\multicolumn{3}{c}{\textbf{Metrics}} \\
\cmidrule(lr){2-3}\cmidrule(lr){4-6}
\textbf{Method} &
\textbf{TFLOPs$\downarrow$} &
\textbf{Speedup rate$\uparrow$} &
\textbf{LPIPS$\downarrow$} &
\textbf{HPS$\uparrow$} &
\textbf{IR DB$\uparrow$} \\
\midrule 

50 steps & 
2991 & $\times$1.00 & 0.000 & 0.306 & 1.008 \\ 

\midrule

\\[-10pt] \midrule

\rowcolor{maincolor}
\multicolumn{6}{c}{\textcolor{emphtextcolor}{\textbf{7 steps}}} \\
\midrule

7 steps of full inference
& 431 & $\times$6.93 &
0.556 & 0.277 & 0.675 \\
\midrule

DiCache
& 477 & $\times$6.27 &
0.546 & 0.276 & 0.758 \\
\rowcolor{black!10} ReCache DiCache
& 477 & $\times$6.27 &
\textbf{0.413} & \textbf{0.296} & \textbf{0.990} \\
\midrule

uniform TaylorSeer ($O=2$)
& 433 & $\times$6.91 &
0.557 & 0.281 & 0.767 \\
\rowcolor{black!10} ReCache TaylorSeer ($O=2$)
& 433 & $\times$6.91 &
\textbf{0.550} & \textbf{0.290} & \textbf{0.976} \\
\midrule

uniform HiCache ($O=2$)
& 433 & $\times$6.91 &
0.557 & 0.278 & 0.689 \\
\rowcolor{black!10} ReCache HiCache ($O=2$)
& 433 & $\times$6.91 &
\textbf{0.498} & \textbf{0.299} & \textbf{1.015} \\
\midrule

DPCache
& 432 & $\times$6.92 &
0.546 & 0.287 & 0.931 \\
\rowcolor{black!10} ReCache Taylor-DP ($O=2$)
& 432 & $\times$6.92 &
\textbf{0.490} & \textbf{0.299} & \textbf{1.012} \\
 \\ [-10pt] 

\midrule
\rowcolor{maincolor}
\multicolumn{6}{c}{\textcolor{emphtextcolor}{\textbf{9 steps}}} \\
\midrule

9 steps of full inference
& 550 & $\times$5.43 &
0.530 & 0.288 & 0.855 \\
\midrule

DiCache
& 594 & $\times$5.04 &
0.456 & 0.296 & 0.974 \\
\rowcolor{black!10} ReCache DiCache
& 594 & $\times$5.04 &
\textbf{0.316} & \textbf{0.302} & \textbf{1.014} \\
\midrule

uniform TaylorSeer ($O=2$)
& 552 & $\times$5.42 &
0.520 & 0.293 & 0.915 \\
\rowcolor{black!10} ReCache TaylorSeer ($O=2$)
& 552 & $\times$5.42 &
\textbf{0.430} & \textbf{0.304} & \textbf{1.008} \\
\midrule

uniform HiCache ($O=2$)
& 552 & $\times$5.42 &
0.521 & 0.290 & 0.886 \\
\rowcolor{black!10} ReCache HiCache ($O=2$)
& 552 & $\times$5.42 &
\textbf{0.331} & \textbf{0.300} & \textbf{0.994} \\
\midrule

DPCache
& 551 & $\times$5.43 &
0.433 & 0.299 & 0.996 \\
\rowcolor{black!10} ReCache Taylor-DP ($O=2$)
& 551 & $\times$5.43 &
\textbf{0.410} & \textbf{0.305} & \textbf{1.004} \\
 \\[-10pt] 

\midrule
\rowcolor{maincolor}
\multicolumn{6}{c}{\textcolor{emphtextcolor}{\textbf{13 steps}}} \\
\midrule

13 steps of full inference
& 789 & $\times$3.79 &
0.484 & 0.295 & 0.931 \\
\midrule

DiCache
& 828 & $\times$3.61 &
0.367 & \textbf{0.303} & 1.001 \\
\rowcolor{black!10} ReCache DiCache
& 828 & $\times$3.61 &
\textbf{0.197} & \textbf{0.303} & \textbf{1.026} \\
\midrule

uniform TaylorSeer ($O=2$)
& 790 & $\times$3.79 &
0.469 & 0.300 & 1.001 \\
\rowcolor{black!10} ReCache TaylorSeer ($O=2$)
& 790 & $\times$3.79 &
\textbf{0.195} & \textbf{0.306} & \textbf{1.041} \\
\midrule

uniform HiCache ($O=2$)
& 790 & $\times$3.79 &
0.471 & 0.299 & 0.967 \\
\rowcolor{black!10} ReCache HiCache ($O=2$)
& 790 & $\times$3.79 &
\textbf{0.212} & \textbf{0.303} & \textbf{1.045} \\
\midrule

DPCache
& 790 & $\times$3.79 &
0.276 & \textbf{0.307} & \textbf{1.046} \\
\rowcolor{black!10} ReCache Taylor-DP ($O=2$)
& 789 & $\times$3.79 &
\textbf{0.244} & 0.306 & 1.034 \\
\bottomrule

\end{tabular}
        }
        \label{tab:main_results_flux}
    \end{minipage}
    \hfill
    \begin{minipage}[t]{0.30\linewidth}
        \vspace{0pt}
        \centering
        \includegraphics[width=1\textwidth]{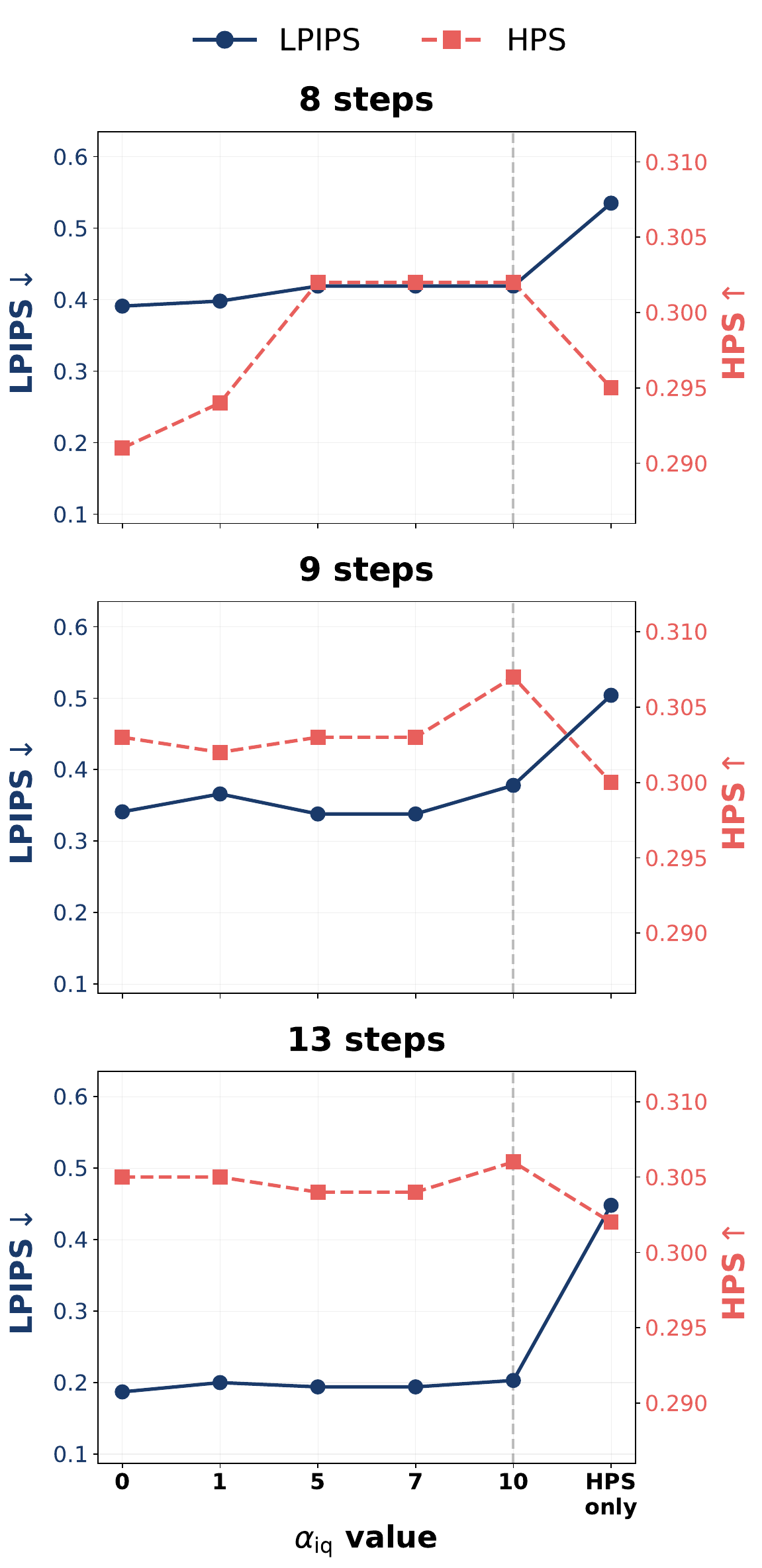}
        \caption{
        Impact of the quality reward coefficient $\losscoef$. Increasing $\alpha_{iq}$ improves HPS but worsens LPIPS, revealing a quality--fidelity trade-off. Optimizing only for HPS causes both metrics to collapse, highlighting the need for the LPIPS fidelity term. We use $\alpha_{iq}=7$ as the best trade-off.
        }
        \label{fig:losscoef_ablation}
    \end{minipage}
    \vspace{-0.5cm}
\end{figure}

\paragraph{Quantitative comparison.}

Tables~\ref{tab:main_results_flux},~\ref{tab:main_results_wan} and~\ref{tab:main_results_hunyuan} report results across all combinations of caching mechanism and budgets $k$ on FLUX.1-dev (image), Wan2.1 and HunyuanVideo (video). Within every (mechanism, budget) cell, ReCache improves over the corresponding hand-designed schedule on both LPIPS fidelity and the perceptual reward metrics (HPSv2 and ImageReward for images; HPSv2 and VBench for video), and the ranking is consistent across all three backbones without exception. Notably, the improvements on video models are also reflected in VBench, showing that ReCache preserves holistic video quality rather than only improving frame-wise metrics. Gains widen as the budget tightens: at $k=13$ the improvements are modest, while at $k=7$---the most aggressive regime, corresponding to roughly $\times 5$--$\times 7$ acceleration---ReCache often closes most of the gap to the full-inference trajectory even where the underlying mechanism's default schedule degrades sharply. This indicates that learning the schedule is a complementary improvement to the underlying caching mechanism rather than a model-specific tuning trick.

\paragraph{Qualitative comparison.}

\begin{figure}[!t]
    \centering
    \includegraphics[width=1\textwidth]{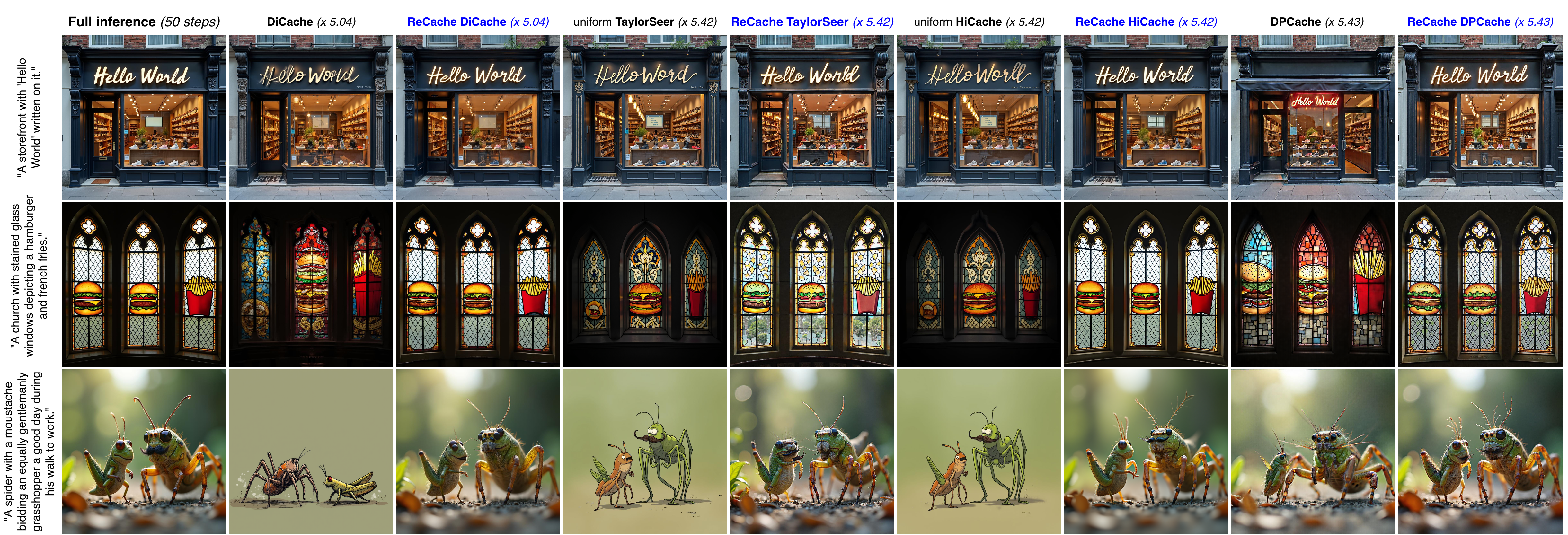}
    \caption{Qualitative comparison on FLUX.1-dev. Under strict compute budgets, the best heuristic baseline schedules (uniform, DiCache, DPCache) fail to maintain image fidelity, corrupting text spelling (top row), object composition (middle row), and fine-grained details (bottom row). By learning an optimal budget-aware schedule, ReCache successfully preserves these critical visual structures at the exact same computational cost.}
    \label{fig:qual_flux}
\end{figure}

Figure~\ref{fig:qual_flux} shows that heuristic schedules can fail under aggressive caching, producing corrupted text, distorted object layouts, and missing fine-grained details. ReCache preserves these visual structures more reliably at the same compute budget, indicating that the learned schedule better identifies the denoising steps where full model evaluation is most critical.

\begin{figure*}[!t]
    \centering

    \begin{minipage}[t]{0.49\textwidth}
        \centering
        \includegraphics[width=\linewidth]{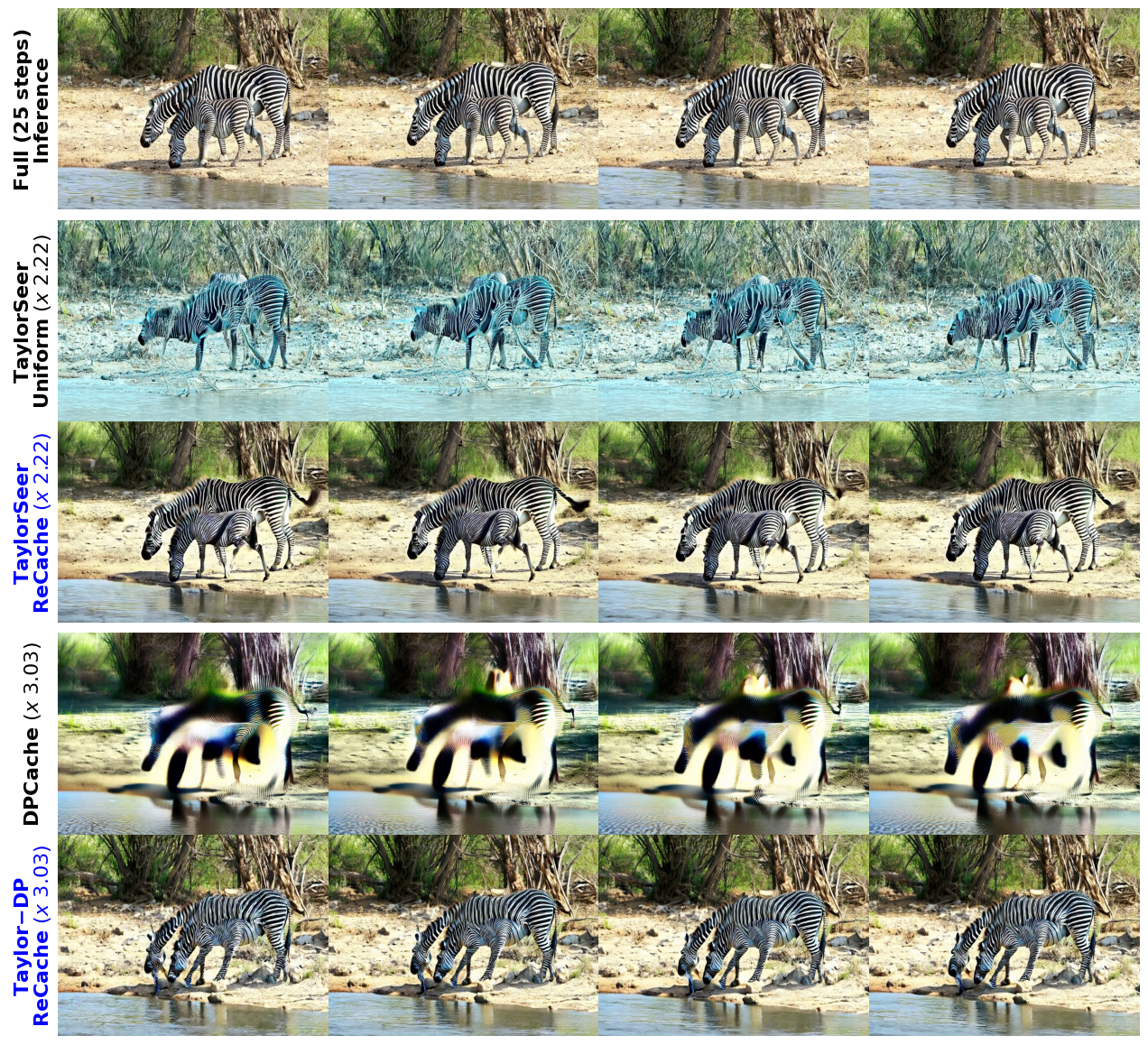}
        \vspace{0.2em}
        \textbf{(a) Wan2.1}
    \end{minipage}
    \hfill
    \begin{minipage}[t]{0.49\textwidth}
        \centering
        \includegraphics[width=\linewidth]{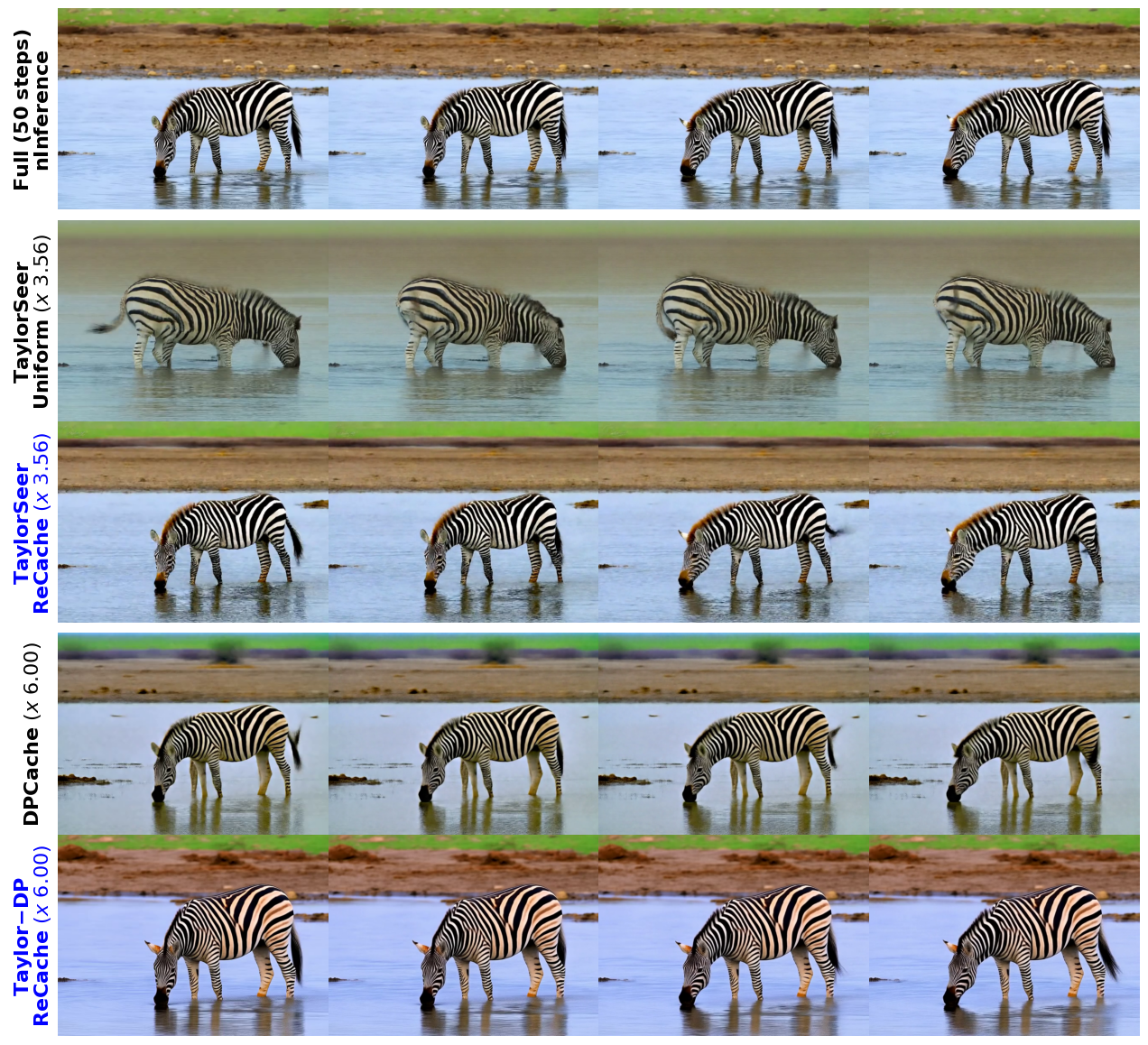}
        \vspace{0.2em}
        \textbf{(b) HunyuanVideo}
    \end{minipage}

    \caption{
        Qualitative comparison on Wan2.1 and HunyuanVideo for the prompt "a zebra bending down to drink water from a river". Under tight computational budgets, baseline scheduling strategies produce visible artifacts such as color degradation and structural blurring, while ReCache, using the same caching mechanism and budget, restores quality and more closely matches full inference.
    }
    \label{fig:qual_video}
\end{figure*}

The same pattern extends to video generation, as shown in Figure~\ref{fig:qual_video}. Under a tight $k=7$ budget, ReCache improves both TaylorSeer and DPCache outputs, preserving sharper spatial details and reducing temporal artifacts across frames.

\begin{table*}[!t]
    \centering

    \begin{minipage}[t]{0.495\textwidth}
        \vspace{0pt}
        \centering
        \captionof{table}{
            Text-to-video results on Wan2.1 across multiple caching budgets. $O$ denotes the forecasting polynomial order.
        }
        \label{tab:main_results_wan}
        \resizebox{\linewidth}{!}{
            \begin{tabular}{l c c c c c}

\toprule
\multirow{2}{*}{\textbf{Method}} &
\multicolumn{2}{c}{\textbf{Acceleration}} &
\multicolumn{3}{c}{\textbf{Metrics}} \\
\cmidrule(lr){2-3}\cmidrule(lr){4-6}
&
\textbf{TFLOPs$\downarrow$} &
\textbf{Speedup$\uparrow$} &
\textbf{LPIPS$\downarrow$} &
\textbf{HPS$\uparrow$} &
\textbf{VBENCH$\uparrow$} \\

\midrule
25 steps & 4139 & $\times$1.00 & 0.000 & 0.234 & 77.717 \\
\midrule

\\[-10pt] 

\midrule
\rowcolor{maincolor}
\multicolumn{6}{c}{\textcolor{emphtextcolor}{\textbf{7 steps}}} \\
\midrule

7 steps of full inference & 1182 & $\times$3.50& 0.506 & 0.188 & 70.652 \\
\midrule

uniform TaylorSeer ($O=1$) & 1184 & $\times$3.50 & 0.499 & 0.185 & 70.430 \\
\rowcolor{black!10} ReCache TaylorSeer ($O=1$) & 1184 & $\times$3.50 & \textbf{0.263} & \textbf{0.228} & \textbf{76.041} \\
\midrule

uniform HiCache ($O=1$) & 1184 & $\times$3.50 & 0.514 & 0.181 & 69.996 \\
\rowcolor{black!10} ReCache HiCache ($O=1$) & 1184 & $\times$3.50 & \textbf{0.266} & \textbf{0.223} & \textbf{75.247} \\
\midrule

uniform Taylor-DP ($O=1$) & 1183 & $\times$3.50 & 0.503 & 0.184 & 70.005 \\
DPCache ($O=1$ & 1183 & $\times$3.50 & 0.420 & 0.201 & 69.643 \\
\rowcolor{black!10} ReCache Taylor-DP ($O=1$) & 1183 & $\times$3.50 & \textbf{0.287} & \textbf{0.229} & \textbf{75.541} \\
 \\ [-10pt] 

\midrule
\rowcolor{maincolor}
\multicolumn{6}{c}{\textcolor{emphtextcolor}{\textbf{13 steps}}} \\
\midrule

13 steps of full inference & 2168 & $\times$1.90 & 0.410 & 0.217 & 74.974 \\
\midrule

uniform TaylorSeer ($O=1$) & 2169 & $\times$1.90 & 0.406 & 0.216 & 75.069 \\
\rowcolor{black!10} ReCache TaylorSeer ($O=1$) & 2169 & $\times$1.90 & \textbf{0.073} & \textbf{0.233} & \textbf{77.708} \\
\midrule

uniform HiCache ($O=1$) & 2169 & $\times$1.90 & 0.413 & 0.214 & 74.612 \\
\rowcolor{black!10} ReCache HiCache ($O=1$) & 2169 & $\times$1.90 & \textbf{0.084} & \textbf{0.232} & \textbf{77.677} \\
\midrule

uniform Taylor-DP ($O=1$) & 2168 & $\times$1.90 & 0.408 & 0.216 & 74.641 \\
DPCache ($O=1$) & 2168 & $\times$1.90 & \textbf{0.074} & 0.233 & 77.816 \\
\rowcolor{black!10} ReCache Taylor-DP ($O=1$) & 2168 & $\times$1.90 & 0.085 & \textbf{0.234} & \textbf{77.893} \\
\bottomrule

\end{tabular}
        }
    \end{minipage}
    \hfill
    \begin{minipage}[t]{0.495\textwidth}
        \vspace{0pt}
        \centering
        \captionof{table}{
            Text-to-video results on HunyuanVideo across multiple caching budgets. $O$ denotes the forecasting polynomial order.
        }
        \label{tab:main_results_hunyuan}
        \resizebox{\linewidth}{!}{
            \begin{tabular}{l c c c c c}

\toprule
\multirow{2}{*}{\textbf{Method}} &
\multicolumn{2}{c}{\textbf{Acceleration}} &
\multicolumn{3}{c}{\textbf{Metrics}} \\
\cmidrule(lr){2-3}\cmidrule(lr){4-6}
&
\textbf{TFLOPs$\downarrow$} &
\textbf{Speedup$\uparrow$} &
\textbf{LPIPS$\downarrow$} &
\textbf{HPS$\uparrow$} &
\textbf{VBENCH$\uparrow$}\\
\midrule

50 steps & 14050 & $\times$1.00 & 0 & 0.248 & 79.309 
\\
\midrule

\\[-10pt] 

\midrule
\rowcolor{maincolor}
\multicolumn{6}{c}{\textcolor{emphtextcolor}{\textbf{7 steps}}} \\ \midrule

7 steps of full inference & 1972 & $\times$7.12 & 0.513 & 0.217 & 75.098 \\
\midrule

uniform TaylorSeer ($O=1$) & 10696 & $\times$1.31 & 0.508 & 0.217 & 76.077 \\
\rowcolor{black!10} ReCache TaylorSeer ($O=1$) & 10696 & $\times$1.31 & \textbf{0.390} & \textbf{0.229} & \textbf{76.264} \\
\midrule

uniform HiCache ($O=1$) & 10696 & $\times$1.31 & 0.512 & 0.212 & 75.383 \\
\rowcolor{black!10} ReCache HiCache ($O=1$) & 10696 & $\times$1.31 & \textbf{0.380} & \textbf{0.226} & \textbf{76.604} \\
\midrule

uniform Taylor-DP ($O=1$) & 1977 & $\times$7.10 & 0.507 & 0.216 & 76.094 \\
DPCache ($O=1$) & 1977 & $\times$7.10 & 0.420 & 0.223 & 74.935 \\
\rowcolor{black!10} ReCache Taylor-DP ($O=1$) & 1977 & $\times$7.10 & \textbf{0.399} & \textbf{0.239} & \textbf{77.396} \\
 \\ [-10pt] 
 
\midrule

\rowcolor{maincolor}
\multicolumn{6}{c}{\textcolor{emphtextcolor}{\textbf{13 steps}}} \\ \midrule

13 steps of full inference & 3657 & $\times$1.31 & 0.428 & 0.231 & 77.708 \\
\midrule

uniform TaylorSeer ($O=1$) & 11164 & $\times$1.26 & 0.417 & 0.236 & 78.219 \\
\rowcolor{black!10} ReCache TaylorSeer ($O=1$) & 11164 & $\times$1.26 & \textbf{0.171} & \textbf{0.241} & \textbf{78.525}\\
\midrule

uniform HiCache ($O=1$) & 11164 & $\times$1.26 & 0.425 & 0.233 & 77.734 \\
\rowcolor{black!10} ReCache HiCache ($O=1$) & 11164 & $\times$1.26 & \textbf{0.168} & \textbf{0.239} & \textbf{78.436} \\
\midrule

uniform Taylor-DP ($O=1$) & 3662 & $\times$1.31 & 0.424 & 0.236 & 78.501 \\
DPCache ($O=1$) & 3662 & $\times$1.31 & 0.188 & \textbf{0.246} & \textbf{79.061} \\
\rowcolor{black!10} ReCache Taylor-DP ($O=1$) & 3662 & $\times$1.31 & \textbf{0.174} & 0.245 & 78.726 \\
\midrule

\end{tabular}

        }
    \end{minipage}
\end{table*}

\subsection{Ablation study}
\label{sec:ablations}

We next analyze the design choices of the proposed ReCache method. All of the ablation experiments are conducted on FLUX.1-dev using the TaylorSeer (O=1) caching mechanism.

\paragraph{Budget-aware logit predictor and nested schedules.}

\begin{wraptable}{t}{0.5\textwidth}
    \centering
    \vspace{-0.5em}
    \caption{Budget-aware ReCache versus per-budget logit predictors on FLUX.1-dev with TaylorSeer.}
    \resizebox{0.48\textwidth}{!}{
        \begin{tabular}{l c c c c}
\toprule
\textbf{Method}
& \multicolumn{2}{c}{\textbf{9 steps}}
& \multicolumn{2}{c}{\textbf{13 steps}} \\
\cmidrule(lr){2-3} \cmidrule(lr){4-5}
& \textbf{LPIPS}$\downarrow$
& \textbf{HPS}$\uparrow$
& \textbf{LPIPS}$\downarrow$
& \textbf{HPS}$\uparrow$ \\
\midrule

single step
& 0.360 & \textbf{0.303}
& 0.220 & \textbf{0.305} \\

budget-aware
& \textbf{0.338} & \textbf{0.303}
& \textbf{0.194} & 0.304 \\

\bottomrule
\end{tabular}
    }
    \label{tab:budget_ablation_main}
\end{wraptable}

ReCache uses a single budget-conditioned MLP that produces logits $\theta = \text{MLP}_\phi(k)$, sharing parameters across budgets. A natural alternative is to train a separate logit predictor for each $k$. Table~\ref{tab:budget_ablation_main} shows that the budget-aware variant improves LPIPS at every evaluated budget and stays within $0.001$ of the per-budget specialists on HPS, all with a single trained model instead of one per budget. We attribute this to joint training, which exposes the MLP to the full range of compute regimes and provides statistical context that a single-budget predictor lacks.

Beyond these metric gains, inspecting the schedules learned by the budget-aware policy reveals a structural property: they are \emph{nested} across all $k \in \{6, \ldots, 15\}$, beyond the evaluated $\{7, 9, 13\}$. The $(k{+}1)$-schedule extends the $k$-schedule by adding a single step rather than reordering its existing selections. Uniform schedules lack this property entirely: their step indices all shift with $k$. This implies that, under the caching objective, the step-importance ranking is largely budget-invariant, and that a single set of logits, independent of $k$, would in principle suffice. Importantly, the optimization does not arrive at this configuration trivially: the MLP produces visibly $k$-dependent logits that sort the inference steps differently across budgets, yet the top-$k$ extractions of those distinct rankings still agree as a nested family. The MLP parameterization does not impose nestedness: it is a property the optimization rediscovers independently at each budget, which we read as evidence that the learned ranking reflects intrinsic structure of the denoising trajectory rather than budget-specific overfitting. Schedule visualizations across budgets are provided in Appendix~\ref{app:further}.

\paragraph{Image-quality reward coefficient.}

The ReCache training objective combines a fidelity term, PLPIPS, with a quality reward, HPS. Figure~\ref{fig:losscoef_ablation} reports the effect of varying the coefficient $\losscoef$, which controls the relative strength of the latter. As shown, for $k \in \{8, 9\}$ steps, increasing $\losscoef$
consistently improves HPS, while LPIPS degrades, suggesting that under strict computational budgets, faithfully replicating the full-inference trajectory becomes difficult, and the model instead learns to generate perceptually pleasing images that score well on human preference. For $k=13$, results remain largely stable across all coefficient values, as sufficient compute allows the predicted schedule to closely follow the full-inference trajectory, which already yields high-quality outputs. We use $\losscoef = 7$ as the best trade-off. Notably, optimizing for HPS alone degrades \emph{both} metrics, which highlights the importance of the fidelity term: it provides strong regularization that anchors the schedule to the full-inference trajectory and prevents the predictor from collapsing to a degenerate solution.

Additional extensive text-to-image results, ablation studies on entropy regularizer and Gumbel-Top-$k$ selection, and analysis of ReCache additional properties like step nestedness of learned schedules and reduced number of the backbone steps can be found in the Appendix~\ref{app:extended_results}.

\section{Conclusion}
\label{sec:conclusion}

In this work, we introduce \textbf{ReCache}, a framework designed to automate the selection of optimal caching schedules for a given inference budget. 
By framing schedule selection as a budget-aware reinforcement learning problem, we move beyond the limitations of hand-crafted heuristics and uniform spacing rules.
Our approach directly optimizes for final output consistency, allowing the model to learn which denoising timesteps are most critical for preserving image and video quality. 
Extensive evaluations across state-of-the-art architectures — including FLUX.1-dev, HunyuanVideo, and Wan2.1 — demonstrate that ReCache consistently discovers schedules that provide a superior quality compared to modern baselines. 

\paragraph{Limitations.} Although ReCache improves existing caching schedules, all feature caching methods provide limited benefits in low-step regimes (1–4 steps), where large gaps between reuse points make cached features less informative.

\paragraph{Broader impacts.} Faster sampling means cheaper generation, which is where most of a deployed diffusion model's lifetime compute actually goes. The flip side of inference efficiency is dual-use: any speedup we give to legitimate users we also give to people generating non-consensual or misleading content. Our method changes the sampling procedure and nothing about the model's safety properties, so it inherits whatever filtering, watermarking, or access controls the underlying model ships with — no better, no worse. We'd encourage anyone deploying it to pair it with provenance tooling appropriate to their setting.

\bibliographystyle{plain}
\bibliography{references}

\newpage
\appendix
\section{Related Work}
\label{app:related_works}

\paragraph{Acceleration of diffusion models}
Diffusion models often require a lot of computationally demanding denoising model evaluations.
There are several ways to alleviate the heavy inference cost and speed up the generative process: one can either make each step faster or reduce the total number of function evaluations.
The former approach includes pruning~\cite{fang2023structuralpruningdiffusionmodels} and quantization techniques~\cite{frantar2022gptq, chee2023quip, xiao2023smoothquant, ashkboos2024quarot, tseng2024quip, dettmers2023spqr}, which are popular in large language modeling and can be adapted for diffusion generation~\cite{li2024svdquant, zhao2024vidit, li2023q, chen2025q, gu2022vector}.
Moreover, computationally heavy attention layers can be sped up using efficient attention~\cite{dao2022flashattention, yuan2024ditfastattn} and token merging~\cite{bolya2023token, saghatchian2025cached} or pruning~\cite{kim2024token, zhang2025training} techniques.
Other papers focus on reducing the number of inference steps to accelerate the generative process by creating an efficient solver~\cite{song2020denoising, karras2022elucidating, lu2022dpm, lu2025dpm, zhao2023unipc, zheng2023dpm} or proposing knowledge distillation~\cite{salimans2022progressive, meng2023distillation}, consistency-based distillation~\cite{song2023consistency, luo2023latent} and different implementations of distribution matching~\cite{sauer2024adversarial, sauer2024fast, yin2024one, yin2024improved, starodubcev2025scale}.
While these architectural and solver-based methods successfully accelerate generation, they are largely orthogonal to caching techniques (such as our proposed method), which focus on avoiding redundant computations inside denoising steps.

\paragraph{Diffusion caching}
Recently, it has been observed that the consequent denoising steps share a lot of similar features, and can therefore be cached and reused in the following steps.
Early papers investigate U-Net~\cite{ronneberger2015u} diffusion model architectures~\cite{rombach2022high, podell2023sdxl, saharia2022photorealistic} and reuse the high-level~\cite{sohl2015deep}, encoder~\cite{sauer2024fast}, or attention~\cite{liu2024faster} features at certain adjacent timesteps.
With the increasing popularity of diffusion transformer-based~\cite{vaswani2017attention} architectures (DiTs)~\cite{peebles2023scalable} in both image~\cite{chen2023pixart, chen2024pixart, blackforest2024flux} and video~\cite{ma2024latte, yang2024cogvideox, wan2025wan, zheng2024open, kong2024hunyuanvideo} domains, research has focused on proposing novel \textit{``What-to-Cache''} strategies~\cite{chen2024delta, selvaraju2024fora, lv2024fastercache}.
\cite{selvaraju2024fora} stores and reuses intermediate outputs from the most computationally demanding layers, while~\cite{chen2024delta} caches the deviations between the outputs of the whole transformer blocks.
Other papers focused on the \textit{``How-to-Use-the-Cache''} question: \cite{bu2025dicache} proposes the dynamic cache trajectory alignment, while \cite{liu2025taylorseer} and \cite{feng2025hicache} incorporate the cache-and-forecast strategy based on the approximation of the higher-order derivatives to better predict the features in future timesteps.
Unlike these approaches that largely rely on fixed heuristic assumptions for caching schedules, ReCache formulates a learned caching policy.

\paragraph{Adaptive caching}
Instead of relying on heuristic assumptions in creating \textit{``When-to-Cache''} schedules, \cite{liu2025timestep, bu2025dicache, kahatapitiya2025adaptive, zhou2025less} create adaptive models that can automatically decide whether the current timestep differs from the cached features and is crucial to recalculate.
To estimate the caching error in real-time and decide whether we should recalculate the current timestep, \cite{kahatapitiya2025adaptive} investigates the residual connections, \cite{zhou2025less} uses the layer inputs, \cite{bu2025dicache} profiles shallow-layer features, while \cite{liu2025timestep} uses a calibrated polynomial.
\cite{zhang2025blockdance} proposes training the lightweight decision-making network to cache the structure-focused blocks of the transformer.
Such adaptive methods show significantly better results, but mostly require inference-time hyperparameter tuning and lack explicit budget control.

\paragraph{ReCache positioning}
To address the limitations of prior work that rely on static schedules and adaptive baselines that require inference-time hyperparameter tuning, ReCache introduces an offline-trained, budget-conditioned paradigm.
By utilizing learned schedule selection, ReCache maximizes generation quality under explicit computational constraints without needing manual real-time adjustments.
\section{Experimental details}
\label{app:details}

\subsection{Modified MS-COCO dataset}
\label{app:modified_mscoco}
As a base dataset of prompts for our text-to-image experiments we created a modified MS-COCO version, where for each of the images we utilized only one caption, which has the biggest length. This ensured we use long and informative prompts with a lot of details to train and evaluate our model on.

\subsection{Baselines implementation} 

FORA~\cite{selvaraju2024fora} is a caching mechanism that caches attention and feedforward layers (FF) in each of the transformer blocks and changes the output to the cached values straightforwardly in the reuse timesteps. $\Delta$-DiT~\cite{chen2024delta} caches the difference between the cache timestep input and the output of the $i$-th transformer block, on the reuse timestep adds this deviation to the current input and evaluates the rest of the blocks as is. DiCache~\cite{bu2025dicache} consists of an Online Probe Profiling Scheme that estimates the caching error in real-time and creates the cache schedule for each sample, and a Dynamic Cache Trajectory Alignment which is a caching mechanism that adaptively approximates the output based on the probe feature trajectory. By a ReCache DiCache we denote that we utilize ReCache policy on the Dynamic Cache Trajectory Alignment. TaylorSeer and HiCache both cache the outputs of the attention and feedforward layers during the cache steps and then predict the current values of a reuse step using the Taylor or Hermite polynomial respectively. DPCache formulates the choice of a caching schedule as a dynamic programming task, it utilizes the TaylorSeer caching mechanism of predicting the values but caches the whole-block outputs instead of separate layers and is called Taylor-DP in its implementation code.

In our HunyuanVideo experiments, we adapt the TaylorSeer baseline to follow the Hugging Face implementation by caching only the attention modules and recomputing the feed-forward layers. This modification is necessary because the original full-cache variant leads to out-of-memory errors in our setting. We use the same feasible TaylorSeer implementation for both the baseline and ReCache TaylorSeer, ensuring a fair comparison. Moreover, the second-order TaylorSeer variant ($O=2$) remains out-of-memory even with attention-only caching; therefore, all video experiments use the first-order variant ($O=1$). This configuration still provides a substantial speed-up, as shown in Table~\ref{tab:main_results_hunyuan}.

\subsection{ReCache training overview}
\label{app:recache_overview}

\paragraph{Policy parameterization.}
We parameterize the schedule logits with a lightweight MLP, $\theta = \text{MLP}_\phi(k)$, that takes the target budget $k$ as input and outputs $N$ logits, one per backbone step. The MLP has approximately $100$K parameters, making its overhead negligible relative to the diffusion backbone $G$. Per-model architecture details (input embedding and hidden dimensions) are reported in the subsections below.

\paragraph{Loss instantiation.}
We instantiate the distance term as patch-wise LPIPS~\cite{zhang2018unreasonable} and the reward as HPSv2~\cite{wu2023human}. The two terms are weighted by $\losscoef$. Entropy regularization uses coefficient $\entropyreg$. Each REINFORCE-LOO update draws $n$ schedules per training example; budgets are sampled from a fixed distribution $q(k)$.

\paragraph{Pre-sampled training set.}
Before training, we generate a fixed dataset of triples---prompt, initial latent $\bx_1^{(i)}$, and the corresponding full-inference output $\xfull^{(i)}$ of $G$. This pre-computation avoids running full inference inside the training loop --- the policy uses the cached $\xfull^{(i)}$ as supervision targets and only pays for cached inference $G(\bx_1^{(i)} \mid \struct, \mathbf{M})$ at each step. We re-generate the dataset once per (model, caching mechanism) pair.

\subsection{Text-to-image with FLUX-1.dev~\cite{blackforest2024flux} implementation details}
\begin{itemize}
    \item Diffusion model:
        \begin{itemize}
            \item Number of inference steps = 50;
            \item Flow Matching Euler Discrete Scheduler;
            \item Image resolution $1024\times1024$;
            \item Guidance scale: 3.5;
        \end{itemize}
    \item Logit predictor model:
        \begin{itemize}
            \item Input embedding dimension = 128;
            \item Hidden dimension = [256, 256, 128];
        \end{itemize}
    \item Dataset:
        \begin{itemize}
            \item Training dataset: 64 prompts from modified MS-COCO;
            \item Test dataset: 1000 prompts from modified MS-COCO;
            \item DrawBench ImageReward dataset: 200 prompts from DrawBench benchmark;
        \end{itemize}
    \item Training configurations:
        \begin{itemize}
            \item Batch size = 1, gradient accumulation = 8;
            \item Number of samples for REINFORCE leave-one-out = 6;
            \item HPS metric in loss with $\losscoef = 7$;
            \item Linear decreasing each gradient step entropy regularizer $\entropyreg = 0.005$;
            \item Number of full inference steps in training is sampled from a weighted distribution with base weight = $0.5$, where $(6, 15)$ steps are sampled with weight 1.0;
            \item 50 epochs of training;
            \item Adam optimizer, learning rate = $0.0005$, betas = $(0.9, 0.999)$, weight decay = $0.0$;
        \end{itemize}
\end{itemize}

\subsection{Text-to-video with HunyuanVideo~\cite{kong2024hunyuanvideo} implementation details}
\begin{itemize}
    \item Diffusion model:
        \begin{itemize}
            \item Number of inference steps = 50;
            \item Flow Matching Euler Discrete Scheduler;
            \item Video resolution $480\times640\times65$;
            \item Guidance scale: 6.0;
        \end{itemize}
    \item Dataset:
        \begin{itemize}
            \item Training dataset: 64 prompts from VBench prompts dataset obtained with stratified sampling from 11 VBench dimensions;
            \item Test dataset: one video for each of 944 prompts from VBench dataset;
        \end{itemize}
    \item Logit predictor model:
        \begin{itemize}
            \item Input embedding dimension = 128;
            \item Hidden dimension = [256, 256, 128];
        \end{itemize}
    \item Training configurations:
        \begin{itemize}
            \item Batch size = 1, gradient accumulation = 8;
            \item Number of samples for REINFORCE leave-one-out = 6;
            \item HPS metric in loss with $\losscoef = 7$;
            \item Linear decreasing each gradient step entropy regularizer $\entropyreg = 0.01$;
            \item Number of full inference steps in training is sampled from a weighted distribution with base weight = $0.5$, where $(6, 15)$ steps are sampled with weight 1.0;
            \item 120 epochs of training;
            \item Adam optimizer, learning rate = $0.03$, betas = $(0.9, 0.999)$, weight decay = $0.0$;
        \end{itemize}
\end{itemize}

\subsection{Text-to-video with Wan2.1~\cite{wan2025wan} implementation details}
\begin{itemize}
    \item Diffusion model:
        \begin{itemize}
            \item Number of inference steps = 25;
            \item Flow Matching Euler Discrete Scheduler;
            \item Video resolution $480\times832\times81$;
            \item Guidance scale: 5.0;
        \end{itemize}
    \item Logit predictor model:
        \begin{itemize}
            \item Input embedding dimension = 128;
            \item Hidden dimension = [256, 256, 128];
        \end{itemize}
    \item Dataset:
        \begin{itemize}
            \item Training dataset: 64 prompts from VBench prompts dataset obtained with stratified sampling from 11 VBench dimensions;
            \item Test dataset: one video for each of 944 prompts from VBench dataset;
        \end{itemize}
    \item Training configurations:
        \begin{itemize}
            \item Batch size = 1, gradient accumulation = 8;
            \item Number of samples for REINFORCE leave-one-out = 6;
            \item HPS metric in loss with $\losscoef = 7$;
            \item Linear decrease for each gradient step entropy regularizer $\entropyreg = 0.02$;
            \item Number of full inference steps in training is sampled from a weighted distribution with base weight = $0.5$, where $(6, 15)$ steps are sampled with weight 1.0;
            \item 120 epochs of training;
            \item Adam optimizer, learning rate = $0.03$, betas = $(0.9, 0.999)$, weight decay = $0.0$;
        \end{itemize}
\end{itemize}
\section{Extended experimental results}
\label{app:extended_results}

In this section, we present additional experimental results, including extended metric evaluations, further analysis of ReCache properties, and additional ablation studies.

\subsection{Extended text-to-image results}

Table~\ref{tab:extented_results_flux} present extended experiments evaluating ReCache on FLUX-1.dev across different caching mechanisms. We compare with both uniform and previously proposed adaptive baseline schedules, where applicable. ImageReward DrawBench (IR DB) metric stand for an ImageReward~\cite{xu2023imagereward} score calculated over 200 DrawBench~\cite{saharia2022photorealistic} prompts. The rest of the metrics -- LPIPS~\cite{zhang2018perceptual}, patch-wise LPIPS (PLPIPS), L1, PSNR, SSIM~\cite{wand2004ssim}, HPSv2~\cite{wu2023human}, CLIP~\cite{hessel2021clipscore}, Aesthetic Quality (AQ)~\cite{schuhmann2022laion}, Image Quality (IQ) and ImageReward (IR) -- are computed on 1000 MS-COCO prompts.

\begin{table}[!t]
    \centering
    \caption{Extended experimental results on FLUX-1.dev text-to-image model. ReCache is evaluated across different caching mechanisms and is compared to uniform and adaptive baselines. The results are grouped based on a computational budget and underlying caching mechanism.}
    \resizebox{\linewidth}{!}{
        \begin{tabular}{l c c c c c c c c c c c c c c c}

\toprule
\multirow{2}{*}{} &
\multicolumn{4}{c}{\textbf{Acceleration}} &
\multicolumn{11}{c}{\textbf{Metrics}} \\
\cmidrule(lr){2-5}\cmidrule(lr){6-16} 
\textbf{\textbf{Method}} &
\textbf{Latency(s)$\downarrow$} &
\textbf{Speed$\uparrow$} &
\textbf{FLOPs(T)$\downarrow$} &
\textbf{Speed$\uparrow$} &
\textbf{LPIPS$\downarrow$} &
\textbf{PLPIPS$\downarrow$} &
\textbf{L1$\downarrow$} &
\textbf{PSNR$\uparrow$} &
\textbf{SSIM$\uparrow$} &
\textbf{HPS$\uparrow$} &
\textbf{CLIP$\uparrow$} &
\textbf{AQ$\uparrow$} &
\textbf{IQ$\uparrow$} &
\textbf{IR$\uparrow$} &
\textbf{IR DB$\uparrow$} \\
\midrule

50 steps
& 10.92 & 1 & 2990.80 & 1 & 
0.0000 & 0.0000 & 0.0000 & $\infty$ & 1.0000 & 0.3063 & 26.98 & 6.49 & 71.21 & 1.1195 & 1.0076 \\ \midrule
 \\ [-10pt] 

\midrule
\rowcolor{maincolor}
\multicolumn{16}{c}{\textcolor{emphtextcolor}{\textbf{7 steps}}} \\
\midrule

7 steps of full inference 
& 1.62 & 6.74 & 431.42 & 6.93 &
0.556 & 0.544 & 0.160 & 13.58 & 0.594 & 0.277 & 27.09 & 6.43 & 65.71 & 0.884 & 0.675 \\

\midrule

uniform FORA ($N=7$)
& 2.63 & 4.15 & 432.54 & 6.91 &  
0.567 & 0.555 & 0.162 & 13.54 & 0.588 & 0.270 & 27.17 & 6.28 & 66.26 & 0.808 & 0.628 \\
\rowcolor{black!10} ReCache FORA
& 2.63 & 4.15 & 432.54 & 6.91 & 
\textbf{0.492} & \textbf{0.474} & \textbf{0.080} & \textbf{18.32} & \textbf{0.665} & \textbf{0.292} & \textbf{27.88} & \textbf{6.38} & \textbf{70.99} & \textbf{1.025} & \textbf{0.981} \\ 

\midrule

uniform $\Delta$-DiT ($N=7$)
& 4.87 & 2.24 & 1285.13 & 2.33 &
0.589 & 0.577 & 0.180 & 12.71 & 0.593 & 0.226 & 25.80 & 5.55 & 45.97 & 0.161 & 0.185 \\ 
TeaCache 
& 4.87 & 2.24 & 1285.13 & 2.33 & 
0.542 & 0.530 & 0.142 & 14.49 & 0.613 & 0.270 & 26.90 & 6.28 & 65.52 & 0.779 & 0.763 \\ 
\rowcolor{black!10} ReCache $\Delta$-DiT
& 4.87 & 2.24 & 1285.13 & 2.33 & 
\textbf{0.452} & \textbf{0.436} & \textbf{0.085} & \textbf{17.74} & \textbf{0.699} & \textbf{0.284} & \textbf{26.96} & \textbf{6.48} & \textbf{67.41} & \textbf{0.964} & \textbf{0.896} \\ 

\midrule

uniform DiCache 
& 1.82 & 6.00 & 477.02 & 6.27 &  
0.553 & 0.541 & 0.160 & 13.58 & 0.597 & 0.277 & 27.05 & 6.34 & 65.92 & 0.889 & 0.658 \\
DiCache 
& 1.82 & 6.00 & 477.02 & 6.27 & 
0.546 & 0.535 & 0.143 & 14.33 & 0.610 & 0.276 & 27.04 & 6.37 & 67.26 & 0.871 & 0.758 \\
\rowcolor{black!10} ReCache DiCache 
& 1.82 & 6.00 & 477.02 & 6.27 &  
\textbf{0.413} & \textbf{0.396} & \textbf{0.069} & \textbf{19.15} & \textbf{0.724} & \textbf{0.296} & \textbf{27.29} & \textbf{6.41} & \textbf{69.90} & \textbf{1.059} & \textbf{0.990} \\

\midrule

uniform TaylorSeer ($O=1$, $N=7$)
& 2.93 & 3.73 & 432.58 & 6.91 & 
0.555 & 0.544 & 0.160 & 13.56 & 0.581 & 0.282 & 27.30 & \textbf{6.38} & 68.39 & 0.934 & 0.780 \\
\rowcolor{black!10} ReCache TaylorSeer ($O=1$)
& 2.93 & 3.73 & 432.58 & 6.91 & 
\textbf{0.499} & \textbf{0.483} & \textbf{0.126} & \textbf{15.07} & \textbf{0.629} & \textbf{0.296} & \textbf{27.54} & 6.34 & \textbf{70.01} & \textbf{1.057} & \textbf{0.972} \\

\midrule

uniform TaylorSeer ($O=2$, $N=7$)
 & 3.09 & 3.53 & 432.61 & 6.91 & 
0.557 & 0.546 & 0.161 & 13.54 & \textbf{0.579} & 0.281 & \textbf{27.28} & \textbf{6.38} & 68.48 & 0.933 & 0.767 \\
\rowcolor{black!10} ReCache TaylorSeer ($O=2$)
 & 3.09 & 3.53 & 432.61 & 6.91 & 
\textbf{0.550} & \textbf{0.539} & \textbf{0.146} & \textbf{14.19} & 0.560 & \textbf{0.290} & 27.25 & 6.31 & \textbf{69.67} & \textbf{1.022} & \textbf{0.976} \\

\midrule

uniform HiCache ($O=1$, $N=7$)
& 3.25 & 3.36 & 432.54 & 6.91 &  
0.557 & 0.545 & 0.160 & 13.58 & 0.589 & 0.278 & 27.28 & 6.34 & 67.57 & 0.888 & 0.688 \\
\rowcolor{black!10} ReCache HiCache ($O=1$)
& 3.25 & 3.36 & 432.54 & 6.91 & 
\textbf{0.441} & \textbf{0.423} & \textbf{0.080} & \textbf{18.27} & \textbf{0.697} & \textbf{0.298} & \textbf{27.64} & \textbf{6.41} & \textbf{70.68} & \textbf{1.052} & \textbf{0.973} \\

\midrule

uniform HiCache ($O=2$, $N=7$)
& 3.47 & 3.15 & 432.54 & 6.91 & 
0.557 & 0.545 & 0.161 & 13.57 & 0.589 & 0.278 & 27.29 & 6.33 & 67.60 & 0.890 & 0.689 \\
\rowcolor{black!10} ReCache HiCache ($O=2$)
& 3.47 & 3.15 & 432.54 & 6.91 & 
\textbf{0.498} & \textbf{0.481} & \textbf{0.106} & \textbf{16.23} & \textbf{0.639} & \textbf{0.299} & \textbf{27.75} & \textbf{6.42} & \textbf{71.02} & \textbf{1.065} & \textbf{1.015} \\

\midrule

uniform Taylor-DP ($O=2$, $N=7$)
& 1.66 & 6.58 & 432.12 & 6.92 & 
0.554 & 0.543 & 0.160 & 13.56 & 0.591 & 0.280 & 27.08 & 6.36 & 67.08 & 0.925 & 0.727 \\
DPCache ($O=2, K=7$)
& 1.66 & 6.58 & 432.12 & 6.92 & 
0.546 & 0.533 & 0.132 & 14.96 & 0.594 & 0.287 & \textbf{27.88} & 6.22 & 69.09 & 1.014 & 0.931 \\
\rowcolor{black!10} ReCache Taylor-DP ($O=2$)
& 1.66 & 6.58 & 432.12 & 6.92 & 
\textbf{0.490} & \textbf{0.477} & \textbf{0.125} & \textbf{15.25} & \textbf{0.630} & \textbf{0.299} & 27.26 & \textbf{6.47} & \textbf{70.31} & \textbf{1.092} & \textbf{1.012} \\
 \\ [-10pt] 

\midrule
\rowcolor{maincolor}
\multicolumn{16}{c}{\textcolor{emphtextcolor}{\textbf{8 steps}}} \\
\midrule

8 steps of full inference 
& 1.84 & 5.93 & 490.94 & 6.09 &
0.541 & 0.529 & 0.155 & 13.77 & 0.602 & 0.283 & 27.11 & 6.44 & 67.21 & 0.940 & 0.754 \\ 

\midrule

uniform FORA ($N=6$)
& 2.82 & 3.87 & 492.03 & 6.08 &  
0.550 & 0.538 & 0.158 & 13.68 & 0.596 & 0.278 & 27.28 & 6.31 & 67.96 & 0.904 & 0.711 \\ 
\rowcolor{black!10} ReCache FORA
& 2.82 & 3.87 & 492.03 & 6.08 & 
\textbf{0.439} & \textbf{0.420} & \textbf{0.076} & \textbf{18.64} & \textbf{0.699} & \textbf{0.297} & \textbf{27.47} & \textbf{6.43} & \textbf{70.83} & \textbf{1.041} & \textbf{0.991} \\

\midrule

uniform $\Delta$-DiT ($N=6$)
& 5.00 & 2.18 & 1324.80 & 2.26 &  
0.575 & 0.563 & 0.175 & 12.89 & 0.599 & 0.241 & 26.11 & 5.82 & 50.58 & 0.374 & 0.333 \\ 
TeaCache 
& 5.00 & 2.18 & 1324.80 & 2.26 & 
0.519 & 0.507 & 0.135 & 14.78 & 0.627 & 0.279 & 27.00 & 6.34 & 67.47 & 0.893 & 0.829 \\ 
\rowcolor{black!10} ReCache $\Delta$-DiT
& 5.00 & 2.18 & 1324.80 & 2.26 & 
\textbf{0.400} & \textbf{0.384} & \textbf{0.067} & \textbf{19.42} & \textbf{0.743} & \textbf{0.289} & \textbf{27.01} & \textbf{6.54} & \textbf{67.79} & \textbf{1.003} & \textbf{0.934} \\ 

\midrule

uniform DiCache 
& 2.03 & 5.38 & 535.48 & 5.59 & 
0.537 & 0.526 & 0.157 & 13.70 & 0.603 & 0.284 & 27.11 & 6.40 & 67.63 & 0.953 & 0.734 \\
DiCache 
& 2.03 & 5.38 & 535.48 & 5.59 &  
0.494 & 0.482 & 0.131 & 14.86 & 0.638 & 0.290 & 26.98 & \textbf{6.48} & 69.60 & 1.012 & 0.897 \\
\rowcolor{black!10} ReCache DiCache 
& 2.03 & 5.38 & 535.48 & 5.59 & 
\textbf{0.392} & \textbf{0.376} & \textbf{0.067} & \textbf{19.28} & \textbf{0.732} & \textbf{0.302} & \textbf{27.24} & 6.45 & \textbf{70.70} & \textbf{1.093} & \textbf{1.005} \\

\midrule

uniform TaylorSeer ($O=1$, $N=6$)
& 3.11 & 3.51 & 492.07 & 6.08 &
0.539 & 0.528 & 0.156 & 13.71 & 0.591 & 0.288 & 27.24 & \textbf{6.41} & 69.47 & 1.002 & 0.825 \\
\rowcolor{black!10} ReCache TaylorSeer ($O=1$)
& 3.11 & 3.51 & 492.07 & 6.08 &
\textbf{0.419} & \textbf{0.403} & \textbf{0.087} & \textbf{17.55} & \textbf{0.693} & \textbf{0.302} & \textbf{27.47} & 6.36 & \textbf{70.88} & \textbf{1.113} & \textbf{0.986} \\

\midrule

uniform TaylorSeer ($O=2$, $N=6$)
& 3.28 & 3.33 & 492.10 & 6.08 &
0.542 & 0.531 & 0.157 & 13.69 & 0.588 & 0.288 & \textbf{27.26} & 6.43 & 69.45 & 1.002 & 0.831 \\
\rowcolor{black!10} ReCache TaylorSeer ($O=2$)
& 3.28 & 3.33 & 492.10 & 6.08 &
\textbf{0.515} & \textbf{0.502} & \textbf{0.139} & \textbf{14.50} & \textbf{0.601} & \textbf{0.297} & 27.15 & \textbf{6.44} & \textbf{69.89} & \textbf{1.058} & \textbf{0.975} \\

\midrule

uniform HiCache ($O=1$, $N=6$)
& 3.44 & 3.17 & 492.03 & 6.08 &  
0.540 & 0.529 & 0.157 & 13.71 & 0.597 & 0.285 & 27.33 & \textbf{6.37} & 68.86 & 0.966 & 0.776 \\
\rowcolor{black!10} ReCache HiCache ($O=1$)
& 3.44 & 3.17 & 492.03 & 6.08 & 
\textbf{0.352} & \textbf{0.339} & \textbf{0.075} & \textbf{18.65} & \textbf{0.738} & \textbf{0.297} & \textbf{27.36} & \textbf{6.37} & \textbf{70.52} & \textbf{1.049} & \textbf{0.984} \\

\midrule

uniform HiCache ($O=2$, $N=6$)
& 3.70 & 2.95 & 492.03 & 6.08 & 
0.540 & 0.529 & 0.157 & 13.71 & 0.597 & 0.285 & 27.31 & 6.36 & 68.85 & 0.966 & 0.773 \\
\rowcolor{black!10} ReCache HiCache ($O=2$)
& 3.70 & 2.95 & 492.03 & 6.08 & 
\textbf{0.432} & \textbf{0.414} & \textbf{0.076} & \textbf{18.61} & \textbf{0.705} & \textbf{0.301} & \textbf{27.81} & \textbf{6.39} & \textbf{70.89} & \textbf{1.081} & \textbf{0.989} \\

\midrule

uniform Taylor-DP ($O=2$, $N=6$)
& 1.87 & 5.84 & 491.63 & 6.08 & 
0.539 & 0.528 & 0.157 & 13.70 & 0.598 & 0.286 & 27.11 & 6.44 & 68.39 & 0.988 & 0.797 \\
DPCache ($O=2, K=8$)
& 1.87 & 5.84 & 491.63 & 6.08 &
0.440 & 0.425 & 0.098 & 16.90 & 0.678 & 0.300 & \textbf{27.36} & 6.41 & 70.35 & 1.102 & 1.030 \\
\rowcolor{black!10} ReCache Taylor-DP ($O=2$)
& 1.87 & 5.84 & 491.63 & 6.08 &
\textbf{0.418} & \textbf{0.405} & \textbf{0.091} & \textbf{17.42} & \textbf{0.688} & \textbf{0.305} & 27.19 & \textbf{6.50} & \textbf{70.59} & \textbf{1.126} & \textbf{1.045} \\
 \\ [-10pt] 

\midrule
\rowcolor{maincolor}
\multicolumn{16}{c}{\textcolor{emphtextcolor}{\textbf{9 steps}}} \\
\midrule

9 steps of full inference 
& 2.06 & 5.30 & 550.47 & 5.43 &
0.530 & 0.518 & 0.150 & 13.98 & 0.609 & 0.288 & 27.05 & 6.47 & 68.11 & 0.983 & 0.855 \\ 

\midrule

uniform FORA ($N=5$)
& 3.01 & 3.63 & 551.53 & 5.42 &  
0.530 & 0.519 & 0.153 & 13.89 & 0.605 & 0.285 & \textbf{27.28} & 6.32 & 69.06 & 0.968 & 0.826 \\ 
\rowcolor{black!10} ReCache FORA
& 3.01 & 3.63 & 551.53 & 5.42 & 
\textbf{0.370} & \textbf{0.355} & \textbf{0.073} & \textbf{18.91} & \textbf{0.729} & \textbf{0.297} & 27.10 & \textbf{6.40} & \textbf{70.45} & \textbf{1.037} & \textbf{0.987} \\ 

\midrule

uniform $\Delta$-DiT ($N=5$)
& 5.15 & 2.12 & 1364.46 & 2.19 & 
0.555 & 0.543 & 0.169 & 13.15 & 0.607 & 0.256 & 26.45 & 6.03 & 56.42 & 0.573 & 0.519 \\ 
TeaCache 
& 5.15 & 2.12 & 1364.46 & 2.19 & 
0.489 & 0.477 & 0.126 & 15.25 & 0.646 & 0.288 & \textbf{27.02} & 6.39 & \textbf{68.98} & 0.984 & 0.876 \\ 
\rowcolor{black!10} ReCache $\Delta$-DiT
& 5.15 & 2.12 & 1364.46 & 2.19 & 
\textbf{0.385} & \textbf{0.369} & \textbf{0.067} & \textbf{19.44} & \textbf{0.748} & \textbf{0.292} & 26.99 & \textbf{6.54} & 68.52 & \textbf{1.019} & \textbf{0.940} \\ 

\midrule

uniform DiCache 
& 2.25 & 4.85 & 593.94 & 5.04 & 
0.519 & 0.508 & 0.151 & 13.93 & 0.613 & 0.289 & 27.04 & 6.41 & 68.84 & 0.998 & 0.845 \\
DiCache 
& 2.25 & 4.85 & 593.94 & 5.04 &
0.456 & 0.445 & 0.117 & 15.56 & 0.663 & 0.296 & 26.92 & \textbf{6.46} & 70.55 & 1.050 & 0.974 \\
\rowcolor{black!10} ReCache DiCache 
& 2.25 & 4.85 & 593.94 & 5.04 &
\textbf{0.316} & \textbf{0.304} & \textbf{0.064} & \textbf{19.61} & \textbf{0.770} & \textbf{0.302} & \textbf{27.08} & 6.45 & \textbf{70.88} & \textbf{1.102} & \textbf{1.014} \\

\midrule

uniform TaylorSeer ($O=1$, $N=5$)
& 3.30 & 3.31 & 551.57 & 5.42 & 
0.518 & 0.507 & 0.150 & 13.94 & 0.605 & 0.293 & 27.12 & \textbf{6.44} & 70.34 & 1.044 & 0.921 \\
\rowcolor{black!10} ReCache TaylorSeer ($O=1$)
& 3.30 & 3.31 & 551.57 & 5.42 & 
\textbf{0.338} & \textbf{0.327} & \textbf{0.082} & \textbf{17.99} & \textbf{0.735} & \textbf{0.303} & \textbf{27.18} & 6.40 & \textbf{71.07} & \textbf{1.117} & \textbf{0.997} \\

\midrule

uniform TaylorSeer ($O=2$, $N=5$)
& 3.49 & 3.13 & 551.60 & 5.42 & 
0.520 & 0.509 & 0.151 & 13.92 & 0.602 & 0.293 & \textbf{27.10} & 6.44 & 70.46 & 1.046 & 0.915 \\
\rowcolor{black!10} ReCache TaylorSeer ($O=2$)
& 3.49 & 3.13 & 551.60 & 5.42 & 
\textbf{0.430} & \textbf{0.417} & \textbf{0.099} & \textbf{16.90} & \textbf{0.667} & \textbf{0.304} & 27.06 & \textbf{6.45} & \textbf{71.09} & \textbf{1.106} & \textbf{1.008} \\

\midrule

uniform HiCache ($O=1$, $N=5$)
& 3.64 & 3.00 & 551.53 & 5.42 &
0.520 & 0.509 & 0.151 & 13.94 & 0.608 & 0.290 & 27.22 & \textbf{6.38} & 69.68 & 1.019 & 0.887 \\
\rowcolor{black!10} ReCache HiCache ($O=1$)
& 3.64 & 3.00 & 551.53 & 5.42 &
\textbf{0.322} & \textbf{0.308} & \textbf{0.062} & \textbf{19.83} & \textbf{0.765} & \textbf{0.296} & \textbf{27.32} & 6.37 & \textbf{70.44} & \textbf{1.055} & \textbf{0.987} \\

\midrule

uniform HiCache ($O=2$, $N=5$)
& 3.89 & 2.81 & 551.53 & 5.42 &
0.521 & 0.510 & 0.151 & 13.93 & 0.607 & 0.290 & 27.22 & \textbf{6.38} & 69.69 & 1.018 & 0.886 \\
\rowcolor{black!10} ReCache HiCache ($O=2$)
& 3.89 & 2.81 & 551.53 & 5.42 &
\textbf{0.331} & \textbf{0.318} & \textbf{0.072} & \textbf{19.04} & \textbf{0.750} & \textbf{0.300} & \textbf{27.35} & 6.34 & \textbf{70.69} & \textbf{1.080} & \textbf{0.994} \\

\midrule

uniform Taylor-DP ($O=2$, $N=5$)
& 2.09 & 5.22 & 551.13 & 5.43 &
0.519 & 0.508 & 0.151 & 13.92 & 0.608 & 0.291 & 27.00 & 6.43 & 69.52 & 1.023 & 0.867 \\
DPCache ($O=2, K=9$)
& 2.09 & 5.22 & 551.13 & 5.43 &
0.433 & 0.417 & \textbf{0.091} & 17.36 & \textbf{0.693} & 0.299 & \textbf{27.58} & 6.37 & 70.21 & 1.112 & 0.996 \\
\rowcolor{black!10} ReCache Taylor-DP ($O=2$)
& 2.09 & 5.22 & 551.13 & 5.43 &
\textbf{0.410} & \textbf{0.398} & \textbf{0.091} & \textbf{17.52} & 0.692 & \textbf{0.305} & 27.20 & \textbf{6.49} & \textbf{70.85} & \textbf{1.133} & \textbf{1.004} \\
 \\ [-10pt] 

\midrule
\rowcolor{maincolor}
\multicolumn{16}{c}{\textcolor{emphtextcolor}{\textbf{13 steps}}} \\
\midrule

13 steps of full inference 
& 2.91 & 3.75 & 788.55 & 3.79 &
0.484 & 0.472 & 0.135 & 14.67 & 0.640 & 0.295 & 27.08 & 6.48 & 69.35 & 1.027 & 0.931 \\ 

\midrule

uniform FORA ($N=3$)
& 3.77 & 2.90 & 789.51 & 3.79 &
0.479 & 0.468 & 0.135 & 14.68 & 0.639 & 0.296 & \textbf{27.19} & 6.38 & 70.12 & 1.055 & 0.940 \\ 
\rowcolor{black!10} ReCache FORA 
& 3.77 & 2.90 & 789.51 & 3.79 &
\textbf{0.256} & \textbf{0.242} & \textbf{0.045} & \textbf{22.32} & \textbf{0.817} & \textbf{0.302} & 27.13 & \textbf{6.40} & \textbf{70.83} & \textbf{1.088} & \textbf{0.994} \\ 

\midrule

uniform $\Delta$-DiT ($N=3$)
& 5.72 & 1.91 & 1523.13 & 1.96 & 
0.505 & 0.493 & 0.149 & 14.00 & 0.634 & 0.283 & 26.91 & 6.39 & 64.83 & 0.925 & 0.790 \\ 
TeaCache 
& 5.72 & 1.91 & 1523.13 & 1.96 &
0.409 & 0.398 & 0.100 & 16.70 & 0.697 & \textbf{0.299} & \textbf{27.07} & 6.43 & \textbf{70.45} & \textbf{1.065} & \textbf{0.996} \\ 
\rowcolor{black!10} ReCache $\Delta$-DiT
& 5.72 & 1.91 & 1523.13 & 1.96 &
\textbf{0.273} & \textbf{0.260} & \textbf{0.046} & \textbf{22.09} & \textbf{0.818} & 0.295 & 26.82 & \textbf{6.53} & 69.46 & 1.046 & 0.975 \\ 

\midrule

uniform DiCache 
& 3.08 & 3.55 & 827.78 & 3.61 &
0.471 & 0.461 & 0.134 & 14.72 & 0.645 & 0.298 & 27.01 & 6.47 & 70.09 & 1.065 & 0.965 \\
DiCache 
& 3.08 & 3.55 & 827.78 & 3.61 &
0.367 & 0.357 & 0.090 & 17.37 & 0.726 & \textbf{0.303} & 26.98 & \textbf{6.48} & \textbf{71.02} & \textbf{1.102} & 1.001 \\
\rowcolor{black!10} ReCache DiCache 
& 3.08 & 3.55 & 827.78 & 3.61 &
\textbf{0.197} & \textbf{0.185} & \textbf{0.031} & \textbf{24.50} & \textbf{0.866} & \textbf{0.303} & \textbf{27.08} & 6.44 & 70.92 & 1.097 & \textbf{1.026} \\

\midrule

uniform TaylorSeer ($O=1$, $N=3$)
& 4.06 & 2.69 & 789.54 & 3.79 &
0.468 & 0.457 & 0.131 & 14.82 & 0.643 & 0.300 & 27.03 & \textbf{6.47} & 70.62 & 1.093 & 0.991 \\
\rowcolor{black!10} ReCache TaylorSeer ($O=1$)
& 4.06 & 2.69 & 789.54 & 3.79 &
\textbf{0.194} & \textbf{0.184} & \textbf{0.038} & \textbf{23.43} & \textbf{0.854} & \textbf{0.304} & \textbf{27.13} & 6.38 & \textbf{71.24} & \textbf{1.115} & \textbf{1.029} \\

\midrule

uniform TaylorSeer ($O=2$, $N=3$)
 & 4.21 & 2.59 & 789.58 & 3.79 &
0.469 & 0.459 & 0.132 & 14.79 & 0.642 & 0.300 & 26.98 & \textbf{6.48} & 70.63 & 1.092 & 1.001 \\
\rowcolor{black!10} ReCache TaylorSeer ($O=2$)
 & 4.21 & 2.59 & 789.58 & 3.79 &
\textbf{0.195} & \textbf{0.186} & \textbf{0.040} & \textbf{22.95} & \textbf{0.853} & \textbf{0.306} & \textbf{27.10} & 6.47 & \textbf{71.34} & \textbf{1.126} & \textbf{1.041} \\

\midrule

uniform HiCache ($O=1$, $N=3$)
& 4.53 & 2.41 & 789.51 & 3.79 &
0.470 & 0.460 & 0.133 & 14.77 & 0.643 & 0.299 & 27.13 & \textbf{6.43} & 70.44 & 1.073 & 0.964 \\
\rowcolor{black!10} ReCache HiCache ($O=1$)
& 4.53 & 2.41 & 789.51 & 3.79 &
\textbf{0.206} & \textbf{0.196} & \textbf{0.040} & \textbf{22.83} & \textbf{0.847} & \textbf{0.303} & \textbf{27.19} & 6.39 & \textbf{71.03} & \textbf{1.097} & \textbf{1.027} \\

\midrule

uniform HiCache ($O=2$, $N=3$)
& 4.64 & 2.35 & 789.51 & 3.79 &
0.471 & 0.460 & 0.133 & 14.76 & 0.642 & 0.299 & 27.12 & \textbf{6.44} & 70.43 & 1.077 & 0.967 \\
\rowcolor{black!10} ReCache HiCache ($O=2$)
& 4.64 & 2.35 & 789.51 & 3.79 &
\textbf{0.212} & \textbf{0.201} & \textbf{0.041} & \textbf{22.63} & \textbf{0.843} & \textbf{0.303} & \textbf{27.22} & 6.39 & \textbf{71.03} & \textbf{1.096} & \textbf{1.045} \\

\midrule

uniform Taylor-DP ($O=2$, $N=3$)
& 2.92 & 3.74 & 789.15 & 3.79 &
0.470 & 0.459 & 0.133 & 14.75 & 0.643 & 0.300 & 27.02 & 6.49 & 70.41 & 1.079 & 0.987 \\
DPCache ($O=2, K=13$)
& 2.92 & 3.74 & 789.15 & 3.79 &
0.276 & 0.264 & 0.053 & 21.03 & 0.804 & \textbf{0.307} & 27.00 & \textbf{6.50} & 71.13 & 1.117 & \textbf{1.046} \\
\rowcolor{black!10} ReCache Taylor-DP ($O=2$, $N=3$)
& 2.92 & 3.74 & 789.15 & 3.79 &
\textbf{0.244} & \textbf{0.230} & \textbf{0.036} & \textbf{23.92} & \textbf{0.845} & 0.306 & \textbf{27.08} & 6.48 & \textbf{71.16} & \textbf{1.123} & 1.034 \\

\midrule

\end{tabular}
    }
    \label{tab:extented_results_flux}
\end{table}

\subsection{Further analysis}
\label{app:further}

The following experiments analyze additional properties of ReCache, including the nested structure of learned schedules, transferability across backbones, and the interaction between caching budgets and the number of backbone inference steps. Unless otherwise stated, all experiments are conducted on the text-to-image generative model FLUX.1-dev with the TaylorSeer $(O=1$) caching strategy.

\begin{figure}[t]
    \centering
    \includegraphics[width=0.98\linewidth]{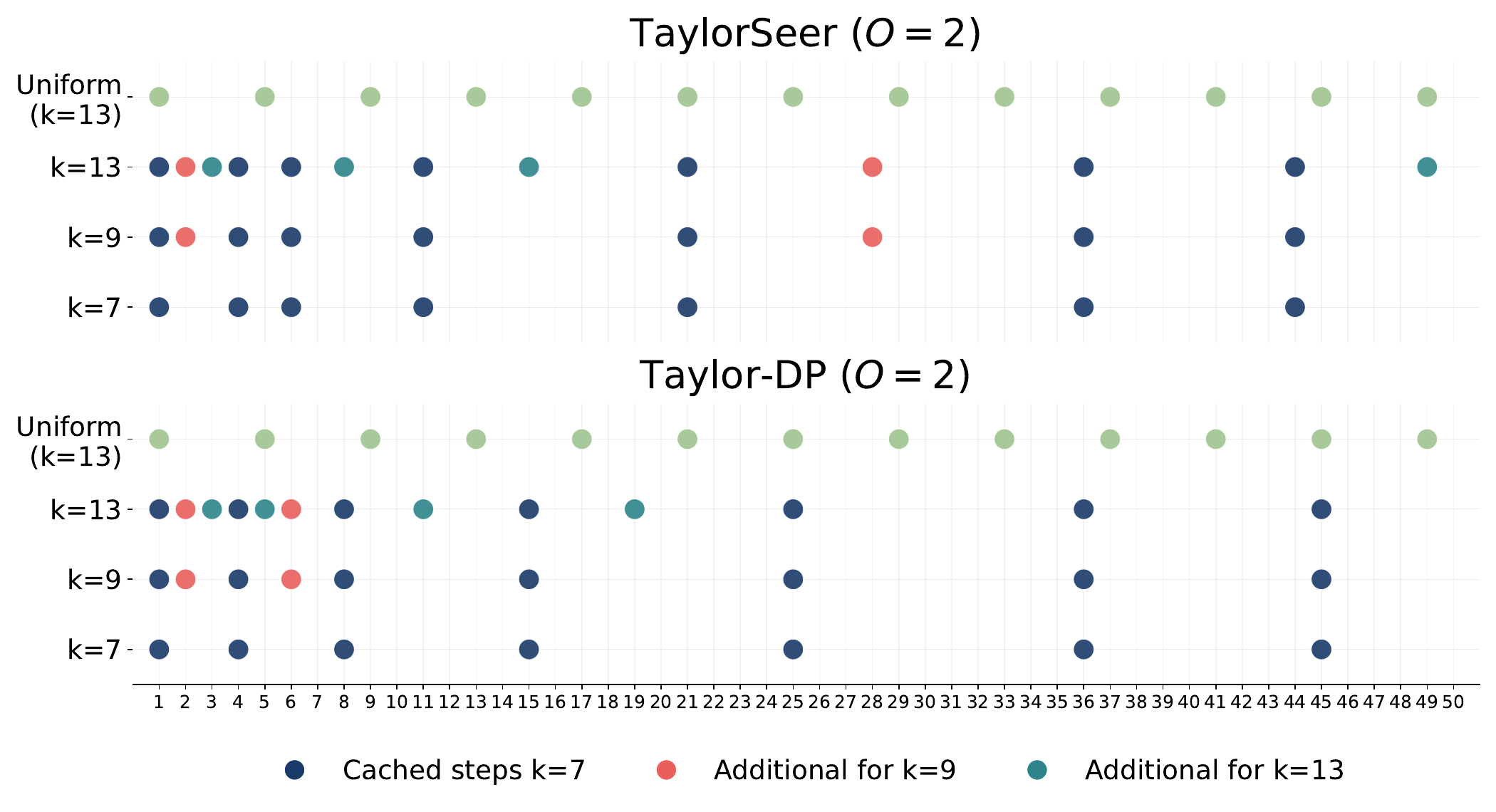}
    \caption{Comparison of cache steps for FLUX.1-dev with the caching mechanisms TaylorSeer~($O=2$) and Taylor-DP~($O=2$)}
    \label{fig:app_cacher_steps}
\end{figure}

\paragraph{Nestedness of learned schedules.}
Figure~\ref{fig:app_cacher_steps} plots the specific cache steps assigned by our budget-aware policy across different target budgets. We illustrate this for FLUX.1-dev using both TaylorSeer $(O=2)$ and Taylor-DP $(O=2)$ mechanisms at $k \in \{7, 9, 13\}$. As the figure shows, the selected steps are strictly additive: the anchor steps identified at the tightest budget ($k=7$) remain completely fixed when the budget is expanded to $k=9$ and $k=13$. The expanded budgets simply activate new steps in the intermediate gaps without shifting the previously selected ones. 

\paragraph{Cross-model schedule transfer.}

\begin{table}[t]
    \centering
    \caption{Cross-model transfer from Wan2.1 to HunyuanVideo. The ReCache schedule for TaylorSeer is learned on Wan2.1 and applied to HunyuanVideo  with a 25-step backbone trajectory without retraining.}
    
    \resizebox{0.67\linewidth}{!}{
        \begin{tabular}{l c c c c c}

\toprule
\multirow{2}{*}{\textbf{Method}} &
\multicolumn{5}{c}{\textbf{Metrics}} \\
\cmidrule(lr){2-6}
&
\textbf{LPIPS$\downarrow$} &
\textbf{PSNR$\uparrow$} &
\textbf{SSIM$\uparrow$} &
\textbf{HPS$\uparrow$} &
\textbf{VBENCH$\uparrow$} \\
\midrule

25 steps & 0 & $\infty$ & 1 & 0.242 & 78.860 \\ 

\midrule

\\[-10pt] \midrule

\rowcolor{maincolor}
\multicolumn{6}{c}{\textcolor{emphtextcolor}{\textbf{7 steps}}} \\
\midrule

7 steps of full inference & 0.462 & 17.75 & 0.690 & 0.211 & 75.098 \\
\midrule

uniform ($O=1$, $N=6$) & 0.451 & 17.64 & 0.684 & 0.219 & 76.131 \\
\rowcolor{black!10} ReCache ($O=1$)& \textbf{0.337} & \textbf{22.93} & \textbf{0.789} & \textbf{0.231} & \textbf{76.807} \\ 

\midrule

\\[-10pt] \midrule

\rowcolor{maincolor}
\multicolumn{6}{c}{\textcolor{emphtextcolor}{\textbf{9 steps}}} \\
\midrule

9 steps of full inference & 0.422 & 18.47 & 0.708 & 0.222 & 75.715 \\
\midrule

uniform ($O=1$, $N=5$) & 0.414 & 18.23 & 0.698 & 0.229 & \textbf{77.795} \\
\rowcolor{black!10} ReCache ($O=1$) & \textbf{0.201} & \textbf{25.06} & \textbf{0.852} & \textbf{0.238} & 77.736 \\ 

\midrule

\\[-10pt] \midrule

\rowcolor{maincolor}
\multicolumn{6}{c}{\textcolor{emphtextcolor}{\textbf{13 steps}}} \\
\midrule

13 steps of full inference & 0.359 & 19.77 & 0.738 & 0.231 & 77.708 \\
\midrule

uniform ($O=1$, $N=3$) & 0.349 & 19.66 & 0.753 & 0.237 & 78.435 \\
\rowcolor{black!10} ReCache ($O=1$) & \textbf{0.078} & \textbf{32.89} & \textbf{0.950} & \textbf{0.242} & \textbf{78.595}\\
\midrule

\end{tabular}

    }
    \label{tab:analysis_wan_to_hunyuan}
\end{table}

A practical question is whether learned caching schedules are specific to a single diffusion backbone or they capture more general structure of the denoising trajectory. To study this, we take a schedule learned by ReCache on Wan2.1 and apply it directly to HunyuanVideo without retraining.

Table~\ref{tab:analysis_wan_to_hunyuan} shows that the transferred schedule remains competitive on HunyuanVideo and substantially improves over standard uniform scheduling. This suggests that effective caching patterns are not purely model-specific. Instead, ReCache appears to identify structural properties of the denoising trajectory that are shared across DiT-based video generation models.

\paragraph{Reducing the number of backbone inference steps.}

So far, the main experiments fix the underlying denoising trajectory length $N$ and reduce the number of full model evaluations $k$ through caching. However, ReCache can also be combined with a shorter backbone trajectory. For example, one can select $K=13$ full evaluations from a $N$=25-step trajectory rather than from a 50-step trajectory. Table~\ref{tab:25_steps_ablation} shows that ReCache still remains effective even with fewer backbone steps. This setting further reduces latency because both cache and reuse evaluations are performed over a shorter denoising trajectory. 

\begin{table}[t]
    \centering
    \caption{
        Impact of reducing the number of backbone steps. Results show that ReCache can be applied on both $N=50$ and $N=25$ to find an optimal schedule.
    }
    \resizebox{0.67\linewidth}{!}{
        \begin{tabular}{l c c c c c c c}
\toprule
\multirow{2}{*}{} & 
\multicolumn{7}{c}{\textbf{Metrics}} \\
\cmidrule(lr){2-8}
\textbf{\textbf{Method}} &
\textbf{LPIPS$\downarrow$} &
\textbf{L1$\downarrow$} & 
\textbf{PSNR$\uparrow$} & 
\textbf{SSIM$\uparrow$} & 
\textbf{HPS$\uparrow$} & 
\textbf{IR$\uparrow$} & 
\textbf{IR DB$\uparrow$} \\
\midrule

50 steps &
0.000 & 0.000 & $\infty$ & 1.000 & 0.306 & 1.120 & 1.008 \\

25 steps &
0.000 & 0.000 & $\infty$ & 1.000 & 0.304 & 1.092 & 1.016  \\
\midrule \\ [-10pt] 

\midrule
\rowcolor{maincolor}
\multicolumn{8}{c}{\textcolor{emphtextcolor}{\textbf{8 steps}}} \\
\midrule

uniform (25 steps) & 
0.611 & 0.146 & 14.58 & 0.543 & 0.224 & 0.046 & 0.015 \\

uniform (50 steps) & 
0.539 & 0.156 & 13.71 & 0.591 & 0.288 & 1.002 & 0.825 \\

\rowcolor{black!10} ReCache (25 steps) & 
0.326 & 0.065 & 19.84 & 0.767 & 0.295 & 1.048 & 0.983 \\ 

\rowcolor{black!10} ReCache (50 steps) & 
0.419 & 0.087 & 17.55 & 0.693 & 0.302 & 1.113 & 0.986 \\
 \\ [-10pt] 

\midrule
\rowcolor{maincolor}
\multicolumn{8}{c}{\textcolor{emphtextcolor}{\textbf{9 steps}}} \\
\midrule

uniform (25 steps) & 
0.564 & 0.142 & 14.72 & 0.590 & 0.246 & 0.496 & 0.328 \\

uniform (50 steps) & 
0.518 & 0.150 & 13.94 & 0.605 & 0.293 & 1.044 & 0.921 \\ 

\rowcolor{black!10} ReCache (25 steps) & 
0.231 & 0.042 & 22.51 & 0.833 & 0.301 & 1.082 & 0.994 \\

\rowcolor{black!10} ReCache (50 steps) & 
0.338 & 0.082 & 17.99 & 0.735 & 0.303 & 1.117 & 0.997 \\
 \\ [-10pt] 

\midrule
\rowcolor{maincolor}
\multicolumn{8}{c}{\textcolor{emphtextcolor}{\textbf{13 steps}}} \\
\midrule

uniform (25 steps) & 
0.512 & 0.130 & 15.18 & 0.628 & 0.281 & 0.922 & 0.730 \\

uniform (50 steps) & 
0.468 & 0.131 & 14.82 & 0.643 & 0.300 & 1.093 & 0.991 \\ 

\rowcolor{black!10} ReCache (25 steps) & 
0.106 & 0.020 & 27.86 & 0.924 & 0.303 & 1.087 & 1.011 \\ 

\rowcolor{black!10} ReCache (50 steps) & 
0.194 & 0.038 & 23.43 & 0.854 & 0.304 & 1.115 & 1.029 \\ 

\bottomrule
\end{tabular}
    }
    \label{tab:25_steps_ablation}
\end{table}

\subsection{Additional ablations}

This section provides extended ablation study on the design choices of ReCache.

\paragraph{Gumbel-Top-$k$ vs.\ Bernoulli schedule selection.}
ReCache uses Gumbel-Top-$k$ to draw a sample of exactly $k$ inference steps, which strictly respects the budget without any auxiliary penalty. An alternative is to use a Bernoulli distribution: each step $i$ is included independently with probability $p_i = \sigma(\theta_i)$ (similar to~\cite{zhang2025blockdance}), and at inference we keep $i$ if $p_i \geq 0.5$. Because this does not guarantee a fixed budget, we add a penalty $\left(\bigl||\struct| - k\bigl|\right) / N$ to the training objective $\mathcal{L}_{\mathrm{reg}}$ with coefficient $\lambda_{\mathrm{bern}}$.

Table~\ref{tab:ablation_gumbel_bern_ablation} shows that the Bernoulli alternative either violates the target budget or requires tuning of $\lambda_{\mathrm{bern}}$ to approximately meet it (e.g., $\lambda_{\mathrm{bern}} = 1$ still violates it). When the budget is roughly matched ($\lambda_{\mathrm{bern}} = 10$), Gumbel-Top-$k$ is more reliable and performs better for $k=9$ and $k=13$. ReCache achieves both strict budget compliance and high-quality results without introducing any budget-related hyperparameter.

\begin{table}[t]
    \centering
    \caption{Comparison of Gumbel-Top-$k$ and Bernoulli selection on FLUX.1-dev for TaylorSeer ($O=1$). Cells where $|\struct| \neq k$ are \colorbox{red!10}{highlighted in red}.}
    \resizebox{0.8\linewidth}{!}{
        \begin{tabular}{l|ccc|ccc|ccc}
\toprule
& \multicolumn{3}{c}{k=7} & \multicolumn{3}{c}{k=9} & \multicolumn{3}{c}{k=13} \\
\cmidrule(lr){2-4}\cmidrule(lr){5-7}\cmidrule(lr){8-10}
& $|s|$  & HPS    & LPIPS & $|s|$  & HPS    & LPIPS & $|s|$  & HPS    & LPIPS  \\
\midrule
Bernoulli $\lambda_{\mathrm{bern}}=1$  & \cellcolor{red!10}15     & \cellcolor{red!10}0.306  & \cellcolor{red!10}0.150 & \cellcolor{red!10}15     & \cellcolor{red!10}0.306  & \cellcolor{red!10}0.150 & \cellcolor{red!10}15     & \cellcolor{red!10}0.306  & \cellcolor{red!10}0.150  \\
Bernoulli $\lambda_{\mathrm{bern}}=10$ & \cellcolor{red!10}8      & \cellcolor{red!10}0.294  & \cellcolor{red!10}0.427 & 9      & \textbf{0.304}  & 0.347 & 13     & 0.303  & 0.242  \\
\rowcolor{black!10} Gumbel-Top‑k                                & 7      & \textbf{0.296}  & \textbf{0.499} & 9      & 0.303  & \textbf{0.338} & 13     & \textbf{0.304}  & \textbf{0.194}  \\
\bottomrule
\end{tabular}
    }
    \label{tab:ablation_gumbel_bern_ablation}
\end{table}

\paragraph{Entropy regularization.}

\begin{figure}[!t]
    \centering
    \includegraphics[width=1\textwidth]{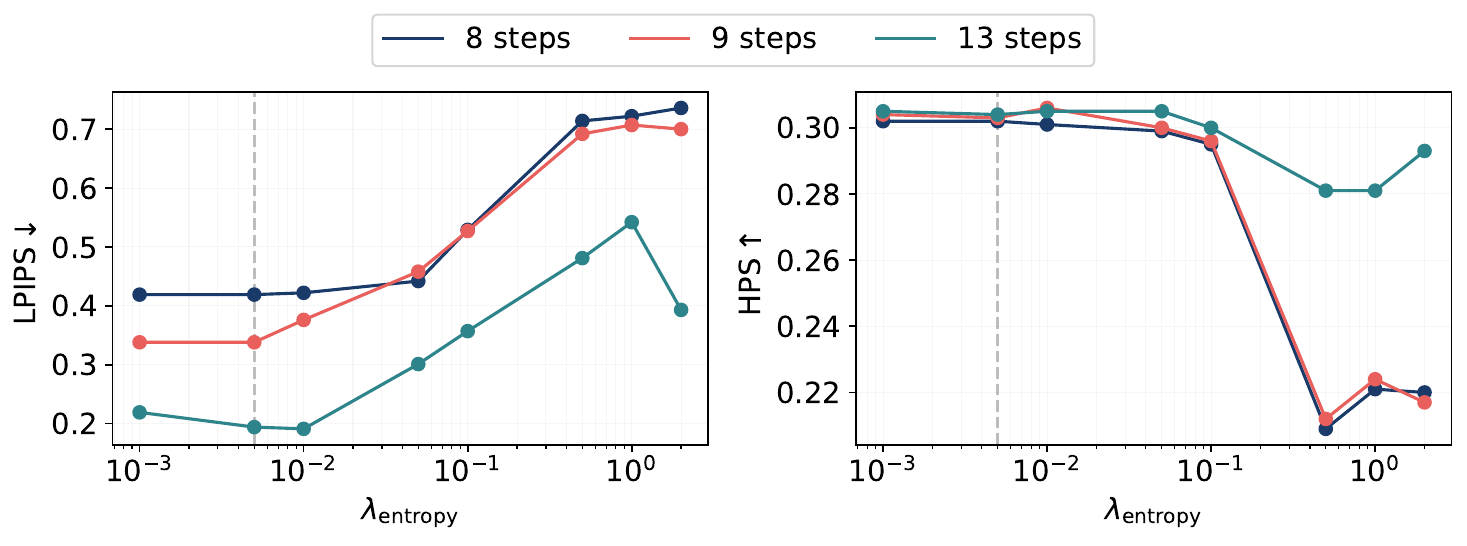}
    \caption{Effect of the entropy regularization coefficient $\entropyreg$ on ReCache. Increasing entropy encourages the discovery of more optimal schedules and is particularly beneficial under looser budget constraints; however, excessively high values can slow convergence.}
    \label{fig:experiments:entropy_coef}
\end{figure}

The entropy regularizer encourages the policy to explore a broader set of caching schedules during training. Figure~\ref{fig:experiments:entropy_coef} ablates the coefficient $\entropyreg$. It is especially helpful for looser budget constraints like $k=13$, where the space of possible schedules is broader and benefits from exploration. The benefit has limits: too-large coefficients slow convergence. In our main experiments we set $\entropyreg = 0.005$ and anneal it to zero, which lets the policy explore freely early in training and converge to a high-quality schedule afterwards.

\paragraph{Image reward loss coefficient.} 

We also provide extended results of the $\losscoef$ values, which controls the relative importance of the HPS term in the training loss. An extensive discussion can be found in Section~\ref{sec:ablations}.

\begin{table*}[t]
    \centering

    \begin{minipage}[t]{0.69\textwidth}
        \vspace{0pt}
        \centering
        \captionof{table}{Loss coefficient $\losscoef$ ablation. The experiments are conducted on FLUX-1.dev with TaylorSeer $(O=1)$ caching mechanisms. $\losscoef=7$ stands for an optimal trade-off between image quality and full-inference fidelity.}
        \resizebox{\linewidth}{!}{
            \begin{tabular}{l c c c c c c c}
\toprule
\multirow{2}{*}{\textbf{Method}} & 
\multicolumn{7}{c}{\textbf{Metrics}} \\
\cmidrule(lr){2-8} & 
\textbf{LPIPS$\downarrow$} &
\textbf{L1$\downarrow$} & 
\textbf{PSNR$\uparrow$} & 
\textbf{SSIM$\uparrow$} & 
\textbf{HPS$\uparrow$} & 
\textbf{IR$\uparrow$} & 
\textbf{IR DB$\uparrow$} \\
\midrule

\rowcolor{maincolor}
\multicolumn{8}{c}{\textcolor{emphtextcolor}{\textbf{8 steps}}} \\ \midrule

uniform & 
0.539 & 0.156 & 13.71 & 0.591 & 0.288 & 1.002 & 0.825 \\ 

ReCache ($\losscoef = 0$) & 
0.391 & 0.086 & 17.28 & 0.700 & 0.291 & 1.065 & 0.973 \\

ReCache ($\losscoef = 1$) & 
0.398 & 0.099 & 16.73 & 0.691 & 0.294 & 1.075 & 0.974 \\

ReCache ($\losscoef = 5$) & 
0.419 & 0.087 & 17.55 & 0.693 & 0.302 & 1.113 & 0.986 \\ 

\rowcolor{black!10} ReCache ($\losscoef = 7$) & 
0.419 & 0.087 & 17.55 & 0.693 & 0.302 & 1.113 & 0.986 \\ 

ReCache ($\losscoef = 10$) & 
0.419 & 0.087 & 17.55 & 0.693 & 0.302 & 1.113 & 0.986 \\ 

ReCache (only HPS) & 
0.535 & 0.127 & 14.99 & 0.603 & 0.295 & 1.062 & 0.968 \\

\midrule

\rowcolor{maincolor}
\multicolumn{8}{c}{\textcolor{emphtextcolor}{\textbf{9 steps}}} \\ \midrule

uniform & 
0.518 & 0.150 & 13.94 & 0.605 & 0.293 & 1.044 & 0.921 \\ 

ReCache ($\losscoef = 0$) & 
0.341 & 0.077 & 18.15 & 0.739 & 0.303 & 1.117 & 1.016 \\

ReCache ($\losscoef = 1$) & 
0.366 & 0.093 & 17.23 & 0.715 & 0.302 & 1.099 & 1.000 \\ 

ReCache ($\losscoef = 5$) & 
0.338 & 0.082 & 17.99 & 0.735 & 0.303 & 1.117 & 0.997 \\ 

\rowcolor{black!10} ReCache ($\losscoef = 7$) & 
0.338 & 0.082 & 17.99 & 0.735 & 0.303 & 1.117 & 0.997 \\ 

ReCache ($\losscoef = 10$) & 
0.378 & 0.082 & 17.99 & 0.718 & 0.307 & 1.131 & 0.992 \\ 

ReCache (only HPS) & 
0.504 & 0.124 & 15.19 & 0.619 & 0.300 & 1.072 & 0.982 \\

\midrule

\rowcolor{maincolor}
\multicolumn{8}{c}{\textcolor{emphtextcolor}{\textbf{13 steps}}} \\ \midrule

uniform & 
0.468 & 0.131 & 14.82 & 0.643 & 0.300 & 1.093 & 0.991 \\ 

ReCache ($\losscoef = 0$) & 
0.187 & 0.035 & 23.94 & 0.864 & 0.305 & 1.115 & 1.031 \\

ReCache ($\losscoef = 1$) & 
0.200 & 0.040 & 22.67 & 0.847 & 0.305 & 1.113 & 1.028 \\ 

ReCache ($\losscoef = 5$) & 
0.194 & 0.038 & 23.43 & 0.854 & 0.304 & 1.115 & 1.029 \\ 

\rowcolor{black!10} ReCache ($\losscoef = 7$) & 
0.194 & 0.038 & 23.43 & 0.854 & 0.304 & 1.115 & 1.029 \\

ReCache ($\losscoef = 10$) & 
0.203 & 0.042 & 22.71 & 0.847 & 0.306 & 1.127 & 1.054 \\

ReCache (HPS only) & 
0.448 & 0.101 & 16.67 & 0.675 & 0.302 & 1.115 & 0.999 \\

\midrule
\end{tabular}
        }
        \label{tab:loss_coef_ablation}
    \end{minipage}
    \hfill
    \begin{minipage}[t]{0.29\textwidth}
        \vspace{0pt}
        \centering
        \captionof{table}{Training time of our ReCache method across different caching mechanisms for both text-to-image and text-to-video models. All our experiments were conducted on H100 GPU.}
        \resizebox{\linewidth}{!}{
           \begin{tabular}{lc}
\toprule
\textbf{Caching Method} & 
\textbf{GPU hours}  \\
 \\ [-10pt] \midrule

\rowcolor{maincolor}
\multicolumn{2}{c}{\textcolor{emphtextcolor}{\textbf{FLUX-1.dev}}} \\

 FORA                 & 20 \\
 $\Delta$-DiT         & 31 \\
 DiCache              & 16 \\
 TaylorSeer ($O=1$)   & 22 \\
 TaylorSeer ($O=2$)   & 23 \\
 HiCache ($O=1$)      & 24 \\
 HiCache ($O=2$)      & 25 \\
 DPTaylorSeer ($O=2$) & 16 \\
 
 \\ [-10pt] \midrule 

\rowcolor{maincolor}
\multicolumn{2}{c}{\textcolor{emphtextcolor}{\textbf{Wan2.1}}} \\ \midrule
 FORA                 & 603 \\
 TaylorSeer ($O=1$)   & 610 \\
 HiCache ($O=1$)      & 613 \\
 DPTaylorSeer ($O=2$) & 588\\
 
 \\ [-10pt] \midrule

\rowcolor{maincolor}
\multicolumn{2}{c}{\textcolor{emphtextcolor}{\textbf{HunyuanVideo}}} \\ \midrule
 TaylorSeer ($O=1$)   & 1110 \\
 HiCache ($O=1$)      & 1112 \\
 DPTaylorSeer ($O=2$) & 937 \\

\bottomrule
\end{tabular}
        }
        \label{tab:training_time}
    \end{minipage}
\end{table*}
\section{Computational resources}
\label{app:computational_resources}

In Table~\ref{tab:training_time} we report the training time of ReCache across different caching mechanisms. FLUX.1-dev is trained for 50 epochs, while the text-to-video models Wan2.1 and HunyuanVideo are trained for 120 epochs. Training time is reported in GPU hours on H100 hardware.


\end{document}